\documentclass[pmlr]{jmlr}% new name PMLR (Proceedings of Machine Learning Research)

\usepackage{longtable}% for long tables

\usepackage{booktabs}
\usepackage[load-configurations=version-1]{siunitx} % newer version
 \usepackage{spverbatim}
 \usepackage{float}
 \usepackage{amsmath}

\usepackage{colortbl} % Add this line to your LaTeX preamble for color support
\usepackage{amsfonts}

\usepackage{etoc}

\theorembodyfont{\upshape}
\theoremheaderfont{\scshape}
\theorempostheader{:}
\theoremsep{\newline}

\jmlrvolume{1}
\jmlryear{2024}
\jmlrworkshop{FM-EduAssess at NeurIPS 2024 Workshop}

\title[]{Automatic Generation of Question Hints for Mathematics Problems using Large Language Models in Educational Technology}

 \author{
   \Name{Junior Cedric Tonga\nametag{\thanks{Work done during internship at EvidenceB.}}}$^{,1,2}$ \Email{juniortonga2022@gmail.com}\\
   \Name{Benjamin Clement}$^2$ \Email{benjamin-c@evidenceb.com}\\
   \Name{Pierre-Yves Oudeyer}$^{2,3}$ \Email{pierre-yves.oudeyer@inria.fr}\\
   \addr $^1$ENS Paris-Saclay, France \\
   $^2$EvidenceB, France \\
   $^3$Inria Bordeaux, France
}

\begin{document}

\maketitle

\begin{abstract}
The automatic generation of hints by Large Language Models (LLMs) within Intelligent Tutoring Systems (ITSs) has shown potential to enhance student learning. However, generating pedagogically sound hints that address student misconceptions and adhere to specific educational objectives remains challenging. This work explores using LLMs (GPT-4o and Llama-3-8B-instruct) as teachers to generate effective hints for students simulated through LLMs (GPT-3.5-turbo, Llama-3-8B-Instruct, or Mistral-7B-instruct-v0.3) tackling math exercises designed for human high-school students, and designed using cognitive science principles. We present here the study of several dimensions: 1) identifying error patterns made by simulated students on secondary-level math exercises; 2) developing various prompts for GPT-4o as a teacher and evaluating their effectiveness in generating hints that enable simulated students to self-correct; and 3) testing the best-performing prompts, based on their ability to produce relevant hints and facilitate error correction, with Llama-3-8B-Instruct as the teacher, allowing for a performance comparison with GPT-4o. The results show that model errors increase with higher temperature settings. Notably, when hints are generated by GPT-4o, the most effective prompts include prompts tailored to specific errors as well as prompts providing general hints based on common mathematical errors. Interestingly, Llama-3-8B-Instruct as a teacher showed better overall performance than GPT-4o. Also the problem-solving and response revision capabilities of the LLMs as students, particularly GPT-3.5-turbo, improved significantly after receiving hints, especially at lower temperature settings. However, models like Mistral-7B-Instruct demonstrated a decline in performance as the temperature increased. This study advances our understanding of the potential and limitations of LLMs in educational contexts, towards integrating these models into pedagogically grounded.
\end{abstract}

\section{Introduction}

Digital education has gained popularity over the last decade, highlighting the importance of Intelligent Tutoring Systems (ITSs). These systems are seen as essential tools to address specific educational challenges, such as the need for personalized learning in a system often reliant on pedagogical teaching and standardized tests, the inaccessibility of private tutoring for everyone, and the difficulty in finding expert tutors at reasonable costs \citep{bray_shadow_1999, graesser_intelligent_2012}. The key feature of ITS is their ability to provide step-by-step guidance to students while they work on problems, with hints playing a crucial role in their capacity to offer this assistance \citep{kinnebrew_integrating_2015}. In the educational context, hints refer to pedagogical questions or suggestions given to learners to help them solve problems, answer questions, or complete tasks. Previous research has shown that providing immediate automated feedback to students within ITS can improve learning outcomes \citep{kochmar_automated_2020, razzaq_effect_2020}.

However, designing such systems remains a challenge. Indeed, a system that directly gives the correct answer when the learner is wrong, which may occurs with Large Language Models (LLMs), does not encourage any effort and can diminish engagement \citep{nie_gpt_2024}. While a system that recognizes the learner's incorrect attempt and provides informative hints related to the learner's existing knowledge encourages critical thinking, problem-solving skills, and independent learning. The challenge to develop such system resides in particular in meeting the diverse learning needs of students and fostering a deeper understanding of complex concepts. These systems can leverage recent advances in Natural Language Processing (NLP), generative AI, and LLMs such as the GPT family models (ChatGPT \citep{chatgpt_introducing_nodate}) or Mistral \citep{jiang2023mistral7b}, Llama \citep{touvron_llama_2023-1}, to be enhanced by integrating LLMs. However, to achieve such a system based on LLMs, these models must meet a wide range of requirements, such as understanding the question and why the student's answer is incorrect, particularly in mathematics, which is the focus of our study, as well as being aligned with educational goals, pedagogical theory, and cognitive processes. By cognitive processes, we mean the skills we aim to develop, the challenges to include in the exercise, and the potential biases in the student's understanding.

\begin{figure}[htbp]
    \centering
    \includegraphics[width=\linewidth]{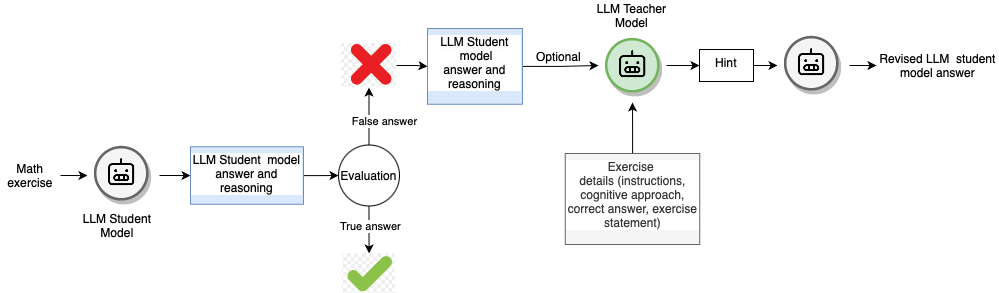}
    \caption{{General approach: An LLM acting as a student solves a math exercise and provides its answer and reasoning. If the answer is incorrect, the exercise details (instructions, cognitive approach, correct answer, exercise statement) are passed, with or without the LLM student's answer and reasoning (depending on the hint-generation prompt type), to another LLM acting as a teacher. The LLM teacher model generates a hint in the form of question using different hint-generation prompts. The hint is then provided to the same LLM student model to revise its response.}}
    \label{fig:overall_approach}
\end{figure}

In this study, we investigate the application of LLMs to generate effective hints for simulated students solving math problems. {Figure \ref{fig:overall_approach} illustrates the overall approach adopted in this paper}. These problems are designed for human students at the high-school level, and are grounded in cognitive science principles \citep{knops2022neural, gros_semantic_2020}. %This research is significant as it may enable the creation of a hint dataset that could be used later on to fine-tune Small LLMs (SLMs). Such SLMs would be valuable for providing guidance to real students, helping them solve problems independently, and could be integrated into ITS. % similar to the one depicted in \ref{fig:its_b}.
Since hint generation involves using LLMs, it is crucial to first understand how these models perform in generating hints when used with simulated students before applying them to real-world scenarios. LLMs can simulate human behaviors, as demonstrated by \citep{markel_gpteach_2023}, who built AI students and studied their interactions with human tutors. Accordingly, we have chosen to use LLMs to simulate students and teacher in our experiments.
The mains contribution of this work are: 1) the evaluation of GPT-4o effectiveness to identify the types of errors made by the student modeled through GPT-3.5-turbo, Llama-3-8B-Instruct, and Mistral-7B-Instruct-v0.3 while solving math exercises based on the temperature parameter; 2) the investigation of the extent to which LLM teachers can generate pedagogically relevant hints i.e. hints that do not provide the answers to simulated students and whether the temperature setting can influence the ability of these simulated student models to self-correct after receiving such hints; 3) the design of several types of prompts, grouped into two categories: specialized prompts and general prompts, used to prompt the teacher model to generate hints -- these two categories differ in their approach: specialized prompts are designed to correct a specific aspect by taking into account the initial answer given by the simulated student, while the general prompts provide hints based on common mistakes that the student model might make when solving math exercises, without considering the simulated student’s initial response; 4) the comparison of these prompts to determine the most effective ones; 5) the evaluation of Llama-3-8B-Instruct as a teacher to generate hints for the GPT-3.5-turbo and Mistral-7B-Instruct-v0.3 students, using the effective prompts identified.

%This paper is structured as follows: Section 2 will present related work, Section 3 will outline our methodology, and Section 4 will detail the experiments, results, and discussions. We will conclude with future work perspectives in Section 5.

\section{Related work}
\subsection{Advances in ITS in Education through LLMs}
Thanks to their ability to provide students with a personalized and effective learning experience, ITSs have gained popularity in the field of education \citep{winkler_unleashing_2018}. According to \citep{feng_systematic_2021}, these systems can be classified into four main categories. \textbf{Dialogue-based tutoring ITS}, such as AutoTutor \citep{graesser_autotutor_2004} and Beetle \citep{dzikovska_beetle_2010} leverage natural language to identify students' misconceptions and respond to their prompts. \textbf{Constraint-based scaffolding models} \citep{mitrovic_effect_2013}, exemplified by KERMIT \citep{suraweera_kermit_2002} use constraints predefined by human experts to respond to student queries. \textbf{Model tracing} \citep{liu_open-ended_2022, sonkar_qdkt_2020} monitors students' knowledge states to capture their problem-solving skills. \textbf{Bayesian network modeling} \citep{corbett_knowledge_1995} extends model tracing by using Bayesian networks.

Furthermore, recent advances in generative artificial intelligence, particularly with the emergence of LLMs such as GPT-4 \citep{bubeck_sparks_2023} from OpenAI and more compact models like Llama \citep{touvron_Llama_2023} from Meta, have demonstrated their potential to significantly enhance these educational technologies. Their remarkable capabilities in generating human-like text and understanding complex linguistic patterns make them particularly well-suited for creating ITS that can interact with students in a more natural and interactive manner.
For example, with Quizbot, \citep{ruan_quizbot_2019} showed the impact of advancements in LLMs on the evolution of educational chatbots. Additionally, \citep{roller_recipes_2021}  proposed a framework to further develop open-domain chatbots. Recently, GPT-4 integrations have been implemented in educational platforms such as Khan Academy’s Khanmigo \citep{khan_meet_nodate} and Quizlet’s Q-Chat \citep{quizlet_q-chat_nodate}, demonstrating the effectiveness of these systems powered by OpenAI's GPT models.
However, most ITSs based on LLMs simply use APIs of these models with a prompt-based strategy, which can limit their scalability while imposing high costs and access restrictions. This is why \citep{sonkar_class_2023} designed a framework called Conversational Learning with Analytical Step-by-Step Strategies (CLASS), aimed at creating ITS powered by performant LLMs, capable of assisting students by posing step-by-step questions. They also introduced a proof-of-concept ITS, called SPOCK, which is trained using the CLASS framework, with a focus on introductory-level college biology content. This framework uses two datasets (Scaffolding dataset and Conversational dataset) generated by GPT-4, which are used to train the SPOCK model (Vicuna 13B) through supervised fine-tuning (SFT). Although some studies address the generation of hints, they primarily focus on the field of biology \citep{sonkar_class_2023}, which is not our area of study. These works are often specific to their domain and are not easily generalizable to other disciplines, such as mathematics. This study to our knowledge, is original as it studies and evaluates the use of LLMs for question hint generation in mathematics, thereby filling a gap in the current literature.

\subsection{LLMs for Feedback and Hint Generation}
 
Before the recent advances in generative artificial intelligence, one of the commonly used approaches for generating feedback or hints in the educational field relied on features designed to detect errors in students' responses. A rule-based system was then employed to provide relevant comments or hints \citep{botelho_leveraging_2023,kochmar_automated_2020,lan_mathematical_2015,razzaq_effect_2020,singh_automated_nodate,song_context-aware_2021}. This approach was popular due to its interpretability and reliability. However, it required significant human effort to adapt to new types of questions.
With the advent of LLMs, a more general approach for generating feedback or hints involves using these advanced models either through prompting \citep{al-hossami_socratic_2023, mcnichols_automated_2024, nguyen_evaluating_2023, steiss_comparing_2023,wang_bridging_2024} or fine-tuning \citep{qinjin_jia_insta-reviewer_2022}. Several studies have been conducted in this area, particularly in the context of programming education. For example, \citep{roest_next-step_2023} explored how LLMs can contribute to programming education by providing students with automated hints for the next steps. They found that most of the feedback messages generated by LLMs describe a specific step to follow and are personalized based on the student’s code and approach. However, these hints can sometimes contain misleading information and lack sufficient detail when students are nearing the end of an exercise.

Similarly, \citep{phung_automating_2023} studied the role of generative AI models in providing human tutor-like hints to help students resolve errors in their faulty programs. However, when prompting pre-trained LLMs, it is crucial that these models exhibit good behavior and a clear understanding of educational objectives. For example, despite these advances, a major challenge these models face is their limited accuracy in handling mathematical calculations. GPT-4, for instance, showed only 59\% accuracy on basic tasks like three-digit multiplication \citep{dziri_faith_2023}. To enhance the mathematical capabilities of LLMs, several methods have been developed, such as the evol-instruct framework of WizardMath \citep{luo_wizardmath_2023}, combining LLMs with symbolic solvers \citep{he-yueya_solving_2023}, or the introduction of “code soliloquies” by \citep{sonkar_code_2023}, which allow for precise invocation of Python calculations whenever a student’s response requires it. 
Nevertheless, the hints or feedback generated by the models should be clear, simple, encouraging, positive in tone, and relevant to the learning objectives \citep{jangra_navigating_2024} as well as address the individual needs of learners \citep{suaib_use_2019, education_recognizing_nodate}.
Many research efforts concentrate on the generation of feedback, which is different from hint generation. These studies are predominantly oriented towards the field of programming, which is distinct from the focus of our paper. Additionally, works in mathematics on feedback or hint generation often benefit from pre-existing datasets, consisting of feedback or hint generated by human experts \citep{scarlatos_improving_2024}. This makes the task of generating feedback somewhat more approachable. In contrast, our study is the first to attempt generating hints in mathematics using LLMs without the advantage of a pre-existing dataset created by human experts. The absence of such reference data makes our approach innovative and underscores the importance of this research, which aims to address this gap in future studies.

\subsection{Simulation of Human Behaviors by LLMs}
 LLMs have the ability to simulate human behaviors, a capability that has shown promising applications in the field of education. For instance, \citep{markel_gpteach_2023} introduced GPTeach, an interactive teacher training tool based on a chat system. This tool allows novice teachers to practice with simulated students, using GPT to take a prompt and generate a response to it \citep{brown_language_2020}. In our study, inspired by the work presented in \citep{markel_gpteach_2023}, we simulate both teachers and students.
Moreover, other studies have suggested that LLMs can be prompted to replicate desired model behaviors \citep{jiang_genline_2021,jiang_promptmaker_2022,liu_pre-train_2021}. Recently, a study \citep{argyle_out_2023} demonstrated that with specific prompting techniques, LLMs can successfully simulate human sub-populations. This work is supported by \citep{arora_ask_2022}, who described various effective prompting techniques. In contrast to these approaches, \citep{park_generative_2023, park_social_2022} used specialized GPT prompting techniques to simulate not just one person, but an entire online community composed of simulated individuals, each with a unique personality.

\section{Methodology}

As a first step, generating high-quality hints requires a rigorous approach to understanding how hints are created when LLMs play the roles of students and teachers. Then, how can we simulate a student and a teacher with LLMs? To achieve this, we used prompt engineering \citep{sahoo_systematic_2024}.

%\subsection{Prompt engineering}
% Prompt engineering is a discipline that encompasses various methods aimed at optimizing the instructions given to an LLM to achieve the best possible results. We used this approach to simulate students and teachers through prompts. A prompt is an instruction or question given to a language model to generate a response or perform a specific task.
% Different prompts were designed for the simulation of students and teachers. Through these simulations, we developed a pipeline to understand the generation of hints by these models.

\subsection{Prompt design for question hint generation}
\label{subsec:comprehension}
To obtain high-quality hints, various approaches were considered for designing prompts and compare them. This led to the development of specific pipelines allowing to test and identify the best prompts for generating relevant and useful hints. A detailed picture of the pipeline can be found in Appendix \ref{fig:pipeline_select}. 
%Figure\ref{fig:pipeline_select} represents the pipeline used to identify best prompts. 
This pipeline is divided into two main stages. The first stage is executed first, followed by the second stage using the datasets obtained from the first stage.

The first stage implements the generation and classification of the student answers. The objective of this step is to create, for each exercise, a diverse dataset of incorrect answers, including different incorrect reasoning or solutions.
    \begin{enumerate}
        \item \textit{Resolution by the student model}: For each exercise, the student model solves it using the prompt described in Appendix \ref{sec:prompts_solve}, producing reasoning and a response.
        \item \textit{Verification of the response by GPT-4o}: The response from the student model is compared to the correct solution using the prompt in Appendix \ref{sec:prompts_check}. \newline
These two steps are repeated a predetermined number of times \textit{num\_simulations} , where \textit{num\_simulations }represents the number of attempts to solve the same exercise by the student model.
    
        \item \textit{Error classification}: 
        After manually evaluating the incorrect responses of simulated students, we identified the following common types of errors: misunderstanding, interpretation, calculation, simplification, algebraic errors, partial answers, term grouping, and incorrect substitution. To automate the determination of these error types, we employed GPT-4o using few-shot prompting \citep{sahoo_systematic_2024}, as detailed in the prompt described in Appendix \ref{sec:prompts_check}, which includes examples for each common error type. GPT-4o  was then used to categorize error types or groups of errors (multiple errors present in a simulated student’s incorrect response), allowing for the creation of a diverse error dataset with various reasoning mistakes or incorrect solutions, using the prompt described in Appendix \ref{sec:prompts_diverse}.
    \end{enumerate}
At the end, a dataset of exercise solutions is produced that includes all the student model’s responses, whether correct or incorrect, corresponding to the number of simulations conducted. We also have an error classification dataset containing various incorrect responses and reasoning by error type.

The second stage implements the generation of hints and revision of various incorrect responses. Only one incorrect answer per error type is retained from the error classification dataset obtained in stage 1 to form the one used in this step.
    \begin{enumerate}
        \item \textit{Hint generation by the teacher Model}: For an  incorrect response in the dataset, the teacher model generates appropriate hints based on the type of prompt used for hint generation. These prompts are described in Appendix \ref{subsec:prompts_hint}.
        \item \textit{Revision of the response by the student model}: The student model uses these hints to revise its initial response using the prompt in the Appendix \ref{sec:prompts_review}.
        \item \textit{Verification of the revised response by GPT-4o}: The revised response is checked again by GPT-4o to ensure it is now correct. This is done by using the prompt described in Appendix \ref{sec:prompts_classif}. \\
        These steps are repeated for each incorrect response in the dataset, as well as for each type of hint generation prompt. The goal is to find the best prompts, i.e., those that allow the best correction of errors by the student models.
    \end{enumerate}
    %At the end of this stage, we obtain a .csv dataset  of hints that contains all the results for the different hint generation prompts used.

The models simulating the students are powered by GPT-3.5-turbo, Llama-3-8B-Instruct, and Mistral-7B-Instruct-v0.3, while those playing the role of the teachers are either GPT-4o or Llama-3-8B-Instruct. GPT-4o was primarily used as the teacher and for intermediate steps in the pipeline, except for the student model stages, because it is currently among the most powerful model available. As mentioned earlier, we defined several prompts. The resolution of exercises and the revision of answers by the student models were done through zero-shot prompting \citep{sahoo_systematic_2024}, meaning without providing examples. We chose to use zero-shot prompting because, when using zero-shot Chain of Thought (CoT) \citep{jin_zero-shot_2024}, the student model made fewer errors, which was not relevant to our study. Indeed, we were looking for a model that produced a balanced mix of erroneous and correct results The phases of answer verification and error classification are also performed by the GPT-4o model in zero-shot prompting. 
%For the phases of answer checking, error classification, and error type determination, we conducted manual verification to ensure the model was not making any mistakes. However, in the vast majority of cases, the results were accurate.
For hint generation, the prompts were also written in zero-shot prompting.
Once we selected the best prompts, a second pipeline was created to evaluate these prompts.

\subsection{Pipeline for evaluating the best prompts}

To evaluate these prompts, we implemented the pipeline illustrated in Figure \ref{fig:prompt_valid}. This pipeline describes a process for evaluating the best prompts. Initially, responses are collected from a dataset of exercise resolutions generated by a student model. If a response is incorrect, a teacher model provides a hint using the best-selected prompt. The student model then revises its response based on this hint. The revised response is subsequently submitted to GPT-4o for re-verification, and the outcome is recorded. This process is repeated for all responses in the dataset. By the end of the pipeline, we have a comprehensive dataset containing all responses before and after hint generation, along with their corresponding evaluations.

\section{Experiments settings}
%We will provide some details on the implementation, dataset and resources used to conduct our experiments.
%\subsection{Dataset}

For these experiments, we worked with the mathematics exercises from MIA Seconde educational software. MIA Seconde is an educational tool developed by EvidenceB, which offers remediation exercises in French and mathematics for students in general, technological, and vocational high school classes. These exercises were initially developed through collaboration with researchers in cognitive science and neuroscience, drawing on insights into how the student’s brain functions and theories about human mathematical cognition. The theory underlying the  MIA Seconde exercises is documented in guides called pedagogical summaries. A pedagogical summary is an official document from EvidenceB that serves as a reference for the design of these exercises. These summaries are organized into different modules, objectives, and activities, each focusing on a specific knowledge or skill.

Modules correspond to an overarching skill developed in the exercises. It generally refers to a sub-discipline of French or mathematics, for example, "quantities and measurements".
Objectives concisely specifies the skills that will be tested. It describes what students should master after practicing the provided exercises, for example, "calculating areas and volumes, and performing unit conversions".
Activities clarify the theme of the exercise. It is characterized by a detailed description of the researchers' cognitive approach and a typical explained exercise. The activity often corresponds to a certain level of difficulty within an objective.

We worked with four different exercises: two exercises for the same module, and two others corresponding to two distinct modules. Each exercise includes its statement, the instructions to follow, the cognitive process, as well as the solution. They are written in French. 
More informations about details on the implementation and the exercises  along with their associated pedagogical elements can be found respectively  in Appendix \ref{subsec:implement} and  \ref{sec:exercises}.

\section{ Experiments and results}
In this section, we analyze the performance of GPT-4o and Llama-3-8B-Instruct as teachers by evaluating their ability to generate helpful hints for solving math problems. We examine student model errors, the best hint generation prompts, and the impact of temperature on the models' problem-solving and revision skills.

\subsection{What types of errors student models make when solving math exercises, and how does it depend on the temperature parameter?}
The types of errors considered are: Comprehension Error: the student does not understand the problem or instructions clearly; Partial Response: the student provides only part of the answer and fails to complete it correctly; Term Grouping Error: the student incorrectly combines or groups terms in an expression; Simplification Error: the student simplifies an expression incorrectly; Calculation Error: the student performs mathematical operations incorrectly; Incorrect Substitution Error: the student substitutes the wrong value in an expression or equation; Interpretation Error: the student misinterprets the instructions or data; Algebraic Error: the student makes mistakes in algebraic manipulations. These errors were identified through manual evaluation and common student mistakes in math, then used in \textit{''few-shot prompting''} with GPT-4o for evaluating student model responses, as mentioned in the error classification phase of the 
pipeline (see \ref{subsec:comprehension}). We manually verified the phases of answer checking, error classification and error type determination to ensure GPT-4o's was not making any mistakes. In most cases, the results were correct (around 98\% of the time).

To analyze the types of errors made by the student models when solving the four exercises under standard settings (default temperature), the first two steps of the initial stage in the pipeline for determining the best prompt for hint generation (see Figure \ref{fig:pipeline_select}) were executed 40 times for each exercise. It is equally interesting to observe how the results vary when the temperature value is adjusted, as temperature is a parameter that controls the creativity and diversity of the responses generated by the model. To explore this, the process was repeated for each temperature value (0, 0.2, 0.5, 0.8, 1) across all models. Studying the effects of different temperature values would also allow us to determine whether the temperature parameter can influence the student model's ability to incorporate a hint during revision process. The verification, evaluation, error detection, and classification steps were performed by GPT-4o, as mentioned earlier, with a temperature value set to 0 to ensure accurate results. It was observed that at higher temperatures ($>$ 0.2), these steps were not reliable.

Studying the effects of different temperature values would also allow us to determine whether the temperature parameter can influence the student model's ability to incorporate a hint during the revision process.

\begin{table}[h!]
\centering
%\vspace{1em}
\caption{A checkmark (\(\checkmark\)) indicates the presence of a specific type of error at a given temperature, while a dash (—) indicates its absence. The table shows errors made by GPT-3.5-turbo (G3.5), Llama-3-8B-Instruct (L8B), and Mistral-7B-Instruct-v0.3 (M7B) in exercise 1 (module 1). As temperature increases, the number and variety of errors tend to rise, varying across models.}
%\vspace{1em}
\resizebox{\textwidth}{!}{
\begin{tabular}{lcccccccccccc}
\hline
Error Type & \multicolumn{3}{c}{Temp 0.2} & \multicolumn{3}{c}{Temp 0.5} & \multicolumn{3}{c}{Temp 0.8} & \multicolumn{3}{c}{Temp 1.0} \\
  & G3.5 & L8B & M7B & G3.5 & L8B & M7B & G3.5 & L8B & M7B & G3.5 & L8B & M7B \\
\hline
Comprehension Error, Grouping of Terms Error & — & — & — & — & — & — & — & - & — & — & — & — \\
\rowcolor[gray]{0.9} % Alternate row color
Interpretation Error, Calculation Error & — & — & — & — & — & — & — & — & — & $\checkmark$ & — & — \\
Comprehension Error, Calculation Error & — & — & — & — & — & — & $\checkmark$ & — & $\checkmark$ & — & — & $\checkmark$ \\
\rowcolor[gray]{0.9} % Alternate row color
Comprehension Error & — & — & $\checkmark$ & $\checkmark$ & — & $\checkmark$ & $\checkmark$ & $\checkmark$ & $\checkmark$ & $\checkmark$ & $\checkmark$ & $\checkmark$ \\
Calculation Error, Interpretation Error & — & — & — & — & — & — & — & — & — & $\checkmark$ & — & — \\
\rowcolor[gray]{0.9} % Alternate row color
Interpretation Error & $\checkmark$ & $\checkmark$ & $\checkmark$ & $\checkmark$ & $\checkmark$ & — & $\checkmark$ & $\checkmark$ & $\checkmark$ & $\checkmark$ & $\checkmark$ & — \\
Grouping of Terms Error & — & — & — & — & — & — & — & — & — & — & - & — \\
\rowcolor[gray]{0.9} % Alternate row color
Calculation Error & $\checkmark$ & — & — & — & — & — & $\checkmark$ & — & — & — & $\checkmark$ & — \\
Simplification Error & — & — & — & $\checkmark$ & — & — & — & — & — & $\checkmark$ & $\checkmark$ & — \\
\rowcolor[gray]{0.9} % Alternate row color
Comprehension Error, Interpretation Error & $\checkmark$ & — & $\checkmark$ & $\checkmark$ & — & $\checkmark$ & — & $\checkmark$ & $\checkmark$ & $\checkmark$ & $\checkmark$ & $\checkmark$ \\
partial response & — & — & — & $\checkmark$ & — & — & $\checkmark$ & — & — & $\checkmark$ & — & — \\
\rowcolor[gray]{0.9}
Simplification Error, Comprehension Error & — & — & — & — & — & — & $\checkmark$ & — & — & $\checkmark$ & — & — \\
Comprehension Error, Calculation Error, Interpretation Error & — & — & — & — & — & — & — & — & $\checkmark$ & — & — & — \\
\hline
\end{tabular}
\label{tab:typeError}
}
\end{table}

We observed that the types of errors made by the student models vary significantly depending on the exercise, the model used, and the temperature value applied. Indeed, the higher the temperature value, the more likely the models are to make errors, as shown in Table \ref{tab:typeError}. This Table summarizes Figure \ref{fig:error_revision_rates_ex1mod1}, offering a concise view of the types of errors encountered at each temperature and for each model in exercise 1 (module 1). In this exercise, Llama made 5 errors, Mistral made 5, and GPT made 9 types of errors. So, Llama-3 and Mistral models tend to make fewer errors types than the GPT-3.5-turbo model. It is also worth mentioning that the Mistral model exhibits a relatively high number of decoding errors compared to the other two. For more details, you can refer to the Appendix \ref{sec:add_res_err}, where we present the evolution of error types based on temperature for other exercises.

 \subsection{What type of prompt is most effective for generating hints with GPT-4o?}
 In order to select the best prompts for generating hints, we defined several prompts, which can be grouped into two categories.

First, \textbf{prompts based on the types of errors made by the student model } aim to correct a key aspect of the student's response. They are based on reasoning, the student's response, the instruction, the cognitive approach, the correct answer, and the exercise, and incorporate these elements into the context. They are labelled as follow: prompt\_hint\_reason is based on the student’s reasoning; %\ref{par:completion_reason}
prompt\_hint\_method is based on the method used by the student; % \ref{par:completion_meth} 
prompt\_hint\_concpt is based on the application of concepts; % \ref{par:completion_concept
prompt\_hint\_calcul is based on calculations; % \ref{par:completion_calcul}
prompt\_hint\_interp is based on problem interpretation; % \ref{par:completion_interp
prompt\_hint\_all combines all the above aspects (calculations, reasoning, interpretation, concepts, method); %\ref{par:completion_all}
prompt\_hint\_part\_res is used to complete partial answers.% \ref{par:completion_part} 
The prompts details are presented in Appendix \ref{subsec:prompts_hint_err}

Second, \textbf{general or baseline prompts} do not consider the student’s reasoning or response, and do not incorporate these elements into the context. They are based only on the exercise, instruction, correct answer, and cognitive approach, incorporating these into the context. They are labelled prompt\_baseline\_one which corresponds to BaselineOne and provides progressive hints to assist the student model %  \ref{par:completion_BOne
and prompt\_baseline\_two which correspond to BaselineTwo prompt which provides hints to assist the student model, based solely on common mistakes that students might make in mathematics. % \ref{par:completion_BTWO}
The prompts details are presented in Appendix \ref{subsec:prompts_hint}

These hints must meet the following criteria:
be in the form of a question, not include the correct answer or a part of it, follow the cognitive approach, be relevant to the exercise. These criteria were introduced into the hint generation prompts to guide the LLM in adhering to these conditions.

The second stage of the pipeline \ref{fig:pipeline_select} is then executed by repeating it 10 times for each of the 9 types of prompts, each temperature value, and each incorrect response from the error classification dataset obtained in stage 1 of this pipeline. Note that when revising the student's response, the same temperature value used during the resolution is applied. For the teacher model, we used a temperature of 1 in all experiments to obtain diverse hints. The mean revision error rate compares prompts by averaging the error revision rates for all encountered errors at a given temperature. For each error type, the error revision rate is calculated as 1 minus the ratio of correct responses to the total number of responses (correct and incorrect) from 10 repetitions.

\begin{table}[h!]
\centering
\caption{Mean revision error rate for each prompt, model, and temperature on exercise 1(module 1). Lower mean revision error rates indicate more effective prompts.}
%\vspace{1em}
\resizebox{0.9\textwidth}{!}{
\begin{tabular}{lcccccccccccc}
 \hline
 Prompt & \multicolumn{3}{c}{Temp 0.2} & \multicolumn{3}{c}{Temp 0.5} & \multicolumn{3}{c}{Temp 0.8} & \multicolumn{3}{c}{Temp 1.0} \\
  & G3.5 & L8B & M7B & G3.5 & L8B & M7B & G3.5 & L8B & M7B & G3.5 & L8B & M7B \\
 \hline
 \rowcolor[gray]{0.9} prompt\_hint\_reason & 0.20 & 0.80 & 0.37 & 0.34 & 0.80 & 0.42 & 0.42 & 0.63 & 0.31 & 0.51 & 0.60 & 0.32 \\
 prompt\_hint\_method & 0.27 & 0.80 & 0.53 & 0.46 & 0.70 & 0.13 & 0.48 & 0.70 & 0.40 & 0.54 & 0.63 & 0.54 \\
 \rowcolor[gray]{0.9} prompt\_hint\_concpt & 0.23 & 0.80 & 0.41 & 0.48 & 0.80 & 0.31 & 0.48 & 0.50 & 0.40 & 0.44 & 0.70 & 0.36 \\
 prompt\_hint\_calcul & 0.27 & 0.90 & 0.44 & 0.44 & 0.90 & 0.38 & 0.50 & 0.70 & 0.53 & 0.43 & 0.55 & 0.11 \\
 \rowcolor[gray]{0.9} prompt\_hint\_interp & 0.20 & 0.60 & 0.37 & 0.40 & 0.90 & 0.30 & 0.42 & 0.73 & 0.39 & 0.44 & 0.65 & 0.78 \\
 prompt\_hint\_all & 0.27 & 0.70 & 0.53 & 0.44 & 0.90 & 0.29 & 0.40 & 0.70 & 0.35 & 0.50 & 0.60 & 0.47 \\
 \rowcolor[gray]{0.9} prompt\_hint\_part\_res & 0.27 & 0.80 & 0.37 & 0.48 & 0.60 & 0.29 & 0.44 & 0.77 & 0.54 & 0.53 & 0.55 & 0.31 \\
 prompt\_baseline\_one & 0.08 & 1.00 & 0.60 & 0.30 & 0.71 & 0.00 & 0.67 & 0.71 & 0.49 & 0.44 & 0.63 & 0.67 \\
 \rowcolor[gray]{0.9} prompt\_baseline\_two & 0.20 & 1.00 & 0.73 & 0.48 & 0.71 & 0.00 & 0.33 & 0.47 & 0.25 & 0.33 & 0.90 & 0.33 \\
 \hline
 \end{tabular}

 \label{tab:meanErrorRate}
 }
 \end{table}

%\begin{table}[h!]
%\centering
%\caption{Rate of amelioration (percentage failure conversion to success) for each model and temperature}
%\resizebox{\textwidth}{!}{
%\begin{tabular}{lcccccccccccc}
%\hline
%Prompt & \multicolumn{3}{c}{Temp 0.2} & \multicolumn{3}{c}{Temp 0.5} & \multicolumn{3}{c}{Temp 0.8} & \multicolumn{3}{c}{Temp 1.0} \\
%\rowcolor[gray]{0.9} & G3.5 & L8B & M7B & G3.5 & L8B & M7B & G3.5 & L8B & M7B & G3.5 & L8B & M7B \\
%\hline
%\rowcolor[gray]{0.9} prompt\_hint\_reason & 0.80 & 0.20 & 0.63 & 0.66 & 0.20 & 0.58 & 0.58 & 0.37 & 0.69 & 0.49 & 0.40 & 0.68 \\
%prompt\_hint\_method & 0.73 & 0.20 & 0.47 & 0.54 & 0.30 & 0.87 & 0.52 & 0.30 & 0.60 & 0.46 & 0.37 & 0.46 \\
%\rowcolor[gray]{0.9} prompt\_hint\_concpt & 0.77 & 0.20 & 0.59 & 0.52 & 0.20 & 0.69 & 0.52 & 0.50 & 0.60 & 0.56 & 0.30 & 0.64 \\
%prompt\_hint\_calcul & 0.73 & 0.10 & 0.56 & 0.56 & 0.10 & 0.63 & 0.50 & 0.30 & 0.47 & 0.57 & 0.45 & 0.89 \\
%\rowcolor[gray]{0.9} prompt\_hint\_interp & 0.80 & 0.40 & 0.63 & 0.60 & 0.10 & 0.70 & 0.58 & 0.27 & 0.61 & 0.56 & 0.35 & 0.22 \\
%prompt\_hint\_all & 0.73 & 0.30 & 0.47 & 0.56 & 0.10 & 0.71 & 0.60 & 0.30 & 0.65 & 0.50 & 0.40 & 0.53 \\
%\rowcolor[gray]{0.9} prompt\_hint\_part\_res & 0.73 & 0.20 & 0.63 & 0.52 & 0.40 & 0.71 & 0.56 & 0.23 & 0.46 & 0.47 & 0.45 & 0.69 \\
%prompt\_baseline\_one & 0.92 & 0.00 & 0.40 & 0.70 & 0.29 & 1.00 & 0.33 & 0.29 & 0.51 & 0.56 & 0.37 & 0.33 \\
%\rowcolor[gray]{0.9} prompt\_baseline\_two & 0.80 & 0.00 & 0.27 & 0.52 & 0.29 & 1.00 & 0.67 & 0.53 & 0.75 & 0.67 & 0.10 & 0.67 \\
%\hline
%\end{tabular}
%\label{tab:meanErrorRate}
%}
%\end{table}

The different prompts enable the learning models to correct their errors, as shown in Table \ref{tab:meanErrorRate}, the lower the mean revision error, the more effective the prompt is. Indeed, regardless of the temperature used, the cues generated by GPT-4o through these prompts allow the LLM models to correct their responses, sometimes entirely. For example, in the case of the Mistral model at a temperature of 0.5, the mean revision error rate is 0 for the baseline-type prompts, which means that all errors where correctly revised. Similarly, for GPT-3.5-turbo, the BaselineTwo prompt has a mean revision error  of 0, indicating that all the cues generated by these prompts enabled GPT-3.5-turbo or Mistral to correct their initial response. Table \ref{tab:meanErrorRate} is a condense version of the Figure \ref{fig:mean_error_ex1mod1} in Appendix \ref{sec:add_res_mean} where more details about these results are described.

%More specifically, Figure\ref{fig:error_rate} shows the revision error rate based on different types of errors. We observe significant corrections for each type of error, regardless of the model used, with particularly marked improvement for GPT-3.5-turbo. However, this model shows difficulties in correcting partial response errors even with the generated cues.

We consider the best prompt to be the one that enables the student model to correct itself the most times over the 10 repetitions. This prompt is identified by selecting the one with the lowest mean revision error rate for each temperature across all exercises. The top prompts are those that appear most frequently as the best. The best specialized prompt was found to be the  \textbf{\textit{one based on calculation errors}}, while the best baseline prompt is \textbf{\textit{BaselineTwo}}. These two prompts were therefore used in the continuation of our experiments.
\subsection{ What is the influence of the temperature parameter on the performance of the student models in solving exercises and revising answers?}

We worked with the best prompts from both categories, specifically the calculation-based prompt for the specialized prompts and BaselineTwo for the baseline-type prompts. To study how temperature could impact the resolution and revision by the student models, we used the validation pipeline shown in Figure \ref{fig:prompt_valid}. Note that when revising the student's response, the same temperature value used during the resolution is applied. Metric such as accuracy was used to quantify the performance of these models using the best prompts. It was computed as the number of correct responses out of the 40 repetitions divided by the number of responses (number of correct+number of incorrect responses).

\begin{figure}[htbp]
    \centering
    \begin{minipage}[b]{0.6\textwidth}
        \centering
        \includegraphics[width=\textwidth]{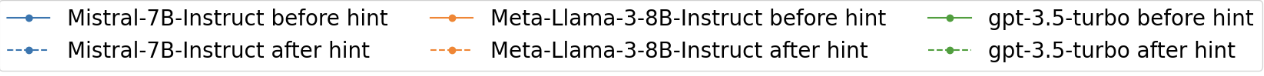}
    \end{minipage}
    \vspace{1em}
    \subfigure[GPT-4o]{
        \centering
        \includegraphics[width=0.45\textwidth]{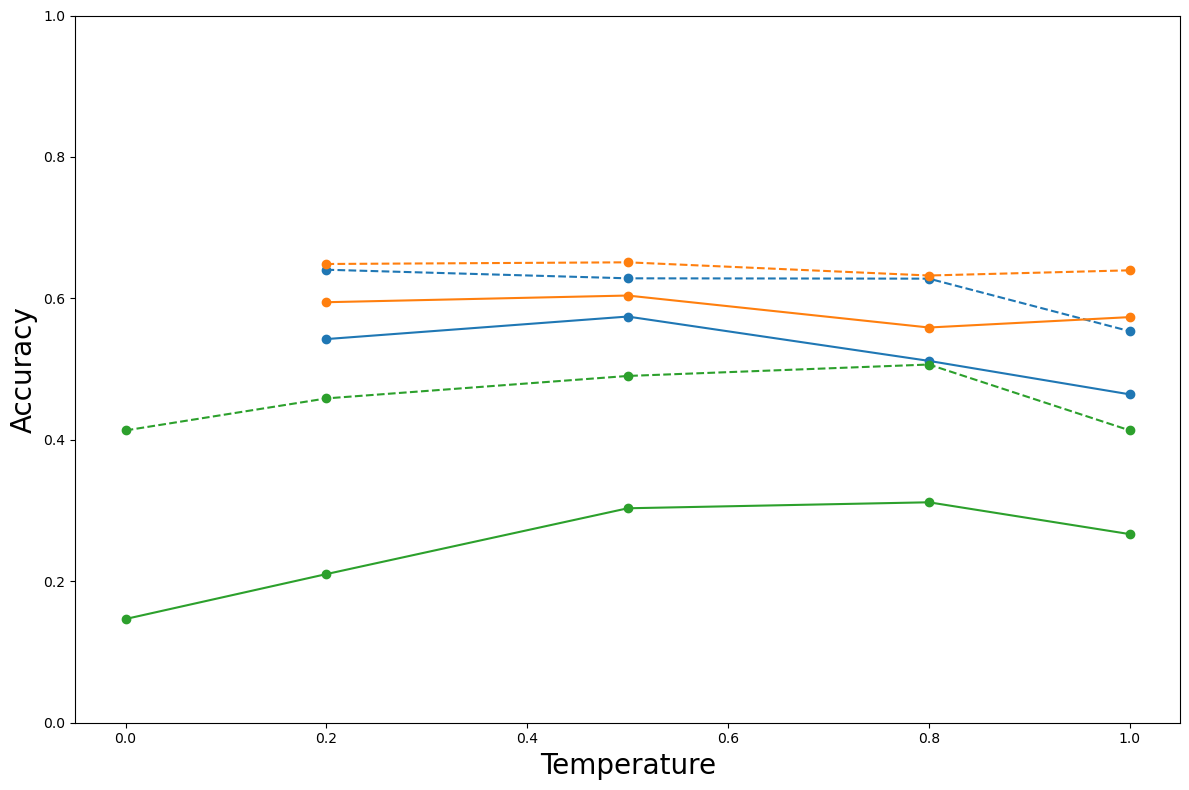}
        \label{fig:gpt_hint_custm}
    }
    \hfill
    \subfigure[Llama-3-8B-instruct]{
        \centering
        \includegraphics[width=0.45\textwidth]{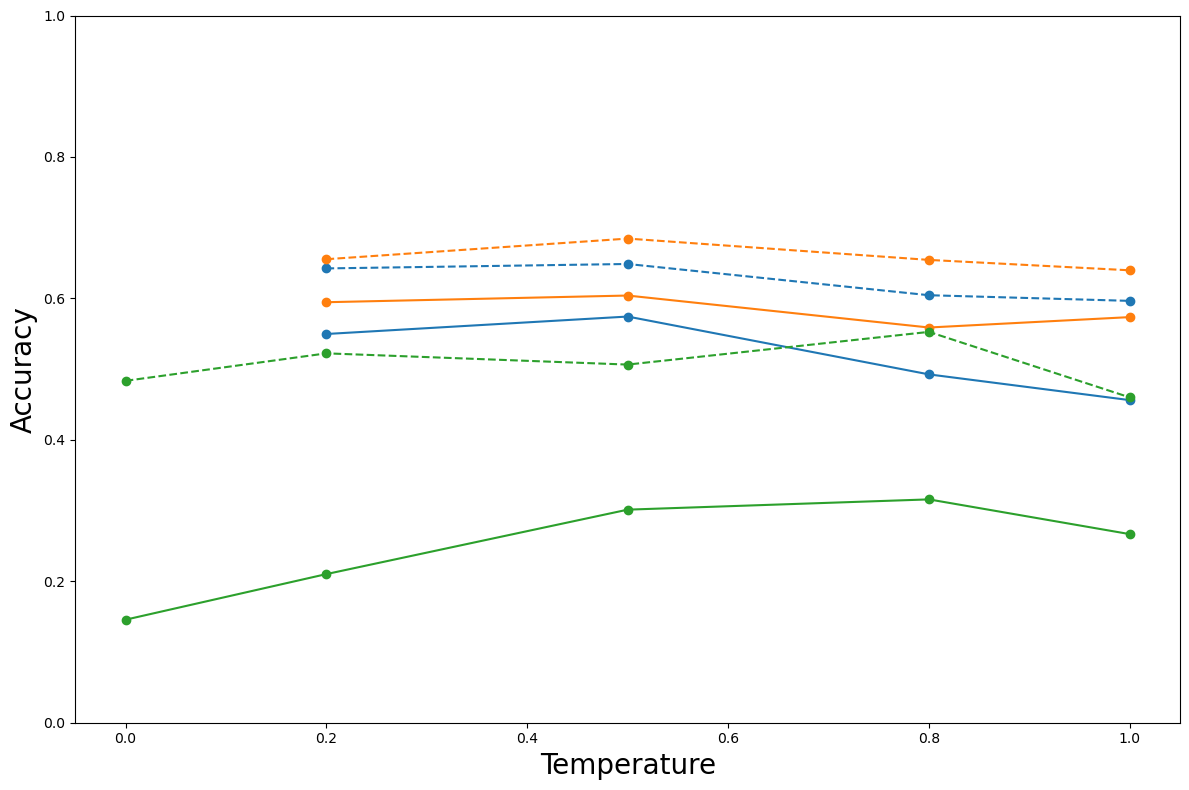}
        \label{fig:Llama_hint_custm}
    }
    \caption{Comparison of accuracy before and after providing hints across four exercises for different student models, using GPT-4o and Llama-3-8B-instruct as teacher models with \textbf{the best specialized hint generation prompt focused on calculation errors}. The results show improved performance when using Llama-3-8B-instruct as the teacher model.}
    \label{fig:custm_hint}
\end{figure}

Figures \ref{fig:open_hint} and \ref{fig:custm_hint} do not show a clear direct link between the ability of student models to solve exercises and revise their answers when GPT-4o or Llama-3-8B-instruct are used as teachers. However, for the GPT student model, we observe that accuracy during both solving and revision decreases when the temperature is set to 1, which is not always the case for the other student models. Conversely, for lower temperatures (e.g., 0 or 0.2), accuracy increases.

\subsection{How does the accuracy of student model problem-solving evolve before and after providing hints when they are guided by Llama-3-8B-Instruct compared to GPT-4o?}

Since our goal is to use LLMs for hint generation, we were curious to see how a smaller language model like Llama-3-8B-Instruct would perform in generating hints. Therefore, we used it as the teacher model in the pipeline shown in Figure \ref{fig:prompt_valid}, utilizing the best prompts.

\begin{figure}[htbp]
    \centering
    \begin{minipage}[b]{0.6\textwidth}
        \centering
        \includegraphics[width=\textwidth]{images/experiments_prompt_validation/Screenshot_2024-09-06_at_13.05.12.png}
    \end{minipage}
    \vspace{1em}
    \subfigure[GPT-4o]{
        \centering
        \includegraphics[width=0.45\textwidth]{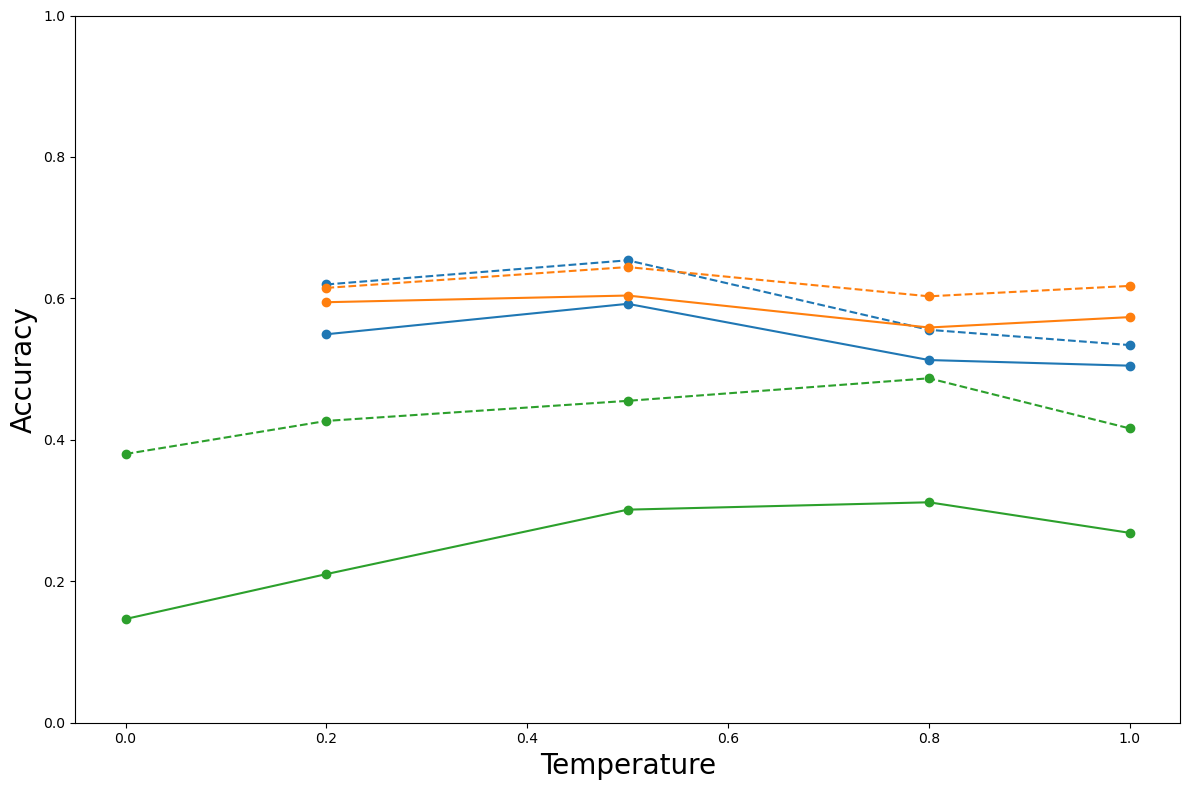}
       % \label{fig:gpt_hint_open}
    }
    \hfill
    \subfigure[Llama-3-8B-instruct]{
        \centering
        \includegraphics[width=0.45\textwidth]{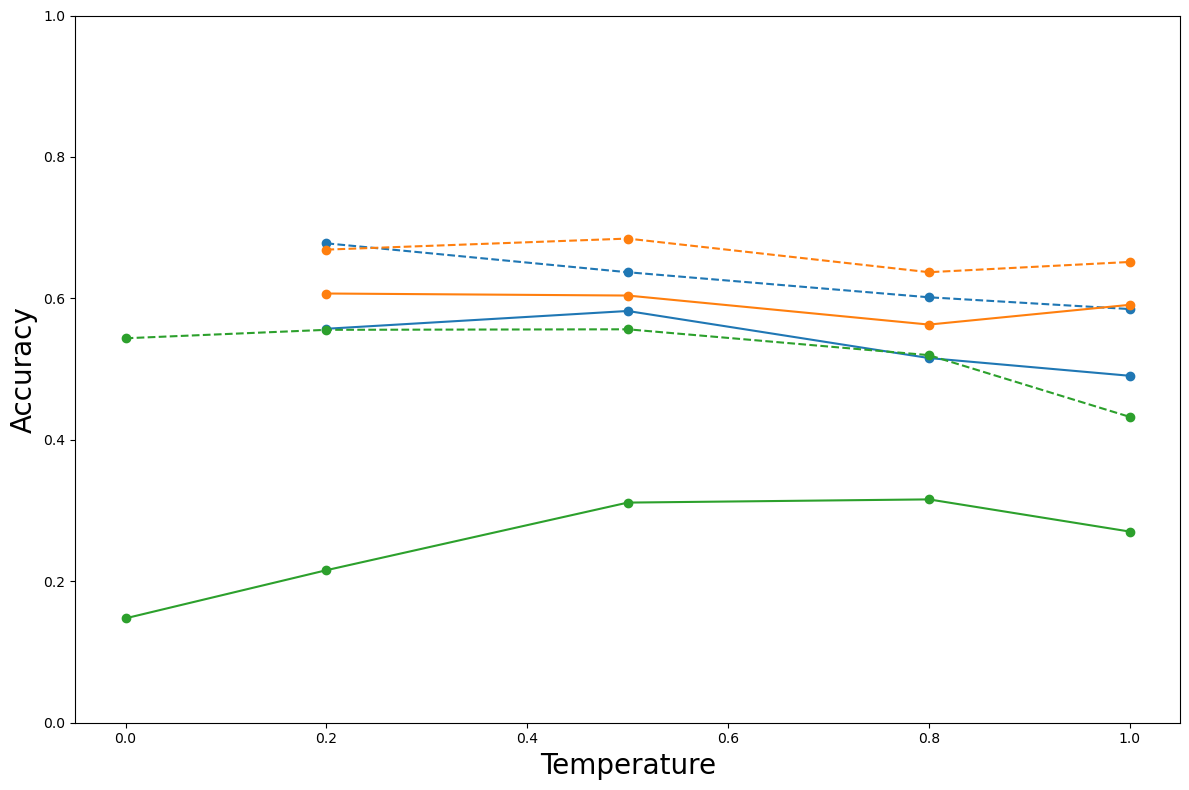}
        %\label{fig:Llama_hint_open}
    }
    \caption{Comparison of accuracy before and after providing hints across four exercises for different student models, using GPT-4o and Llama-3-8B-instruct as teacher models with \textbf{the best baseline-type hint generation prompt, named BaselineTwo}. The results show improved performance when using Llama-3-8B-instruct as the teacher model.}
    \label{fig:open_hint}
\end{figure}

Figures \ref{fig:open_hint} and \ref{fig:custm_hint} show that whether the BaselineTwo prompt or the one based on calculation errors is used with GPT-4o or Llama-3-8B-instruct, the models manage to correct themselves. A notable improvement is particularly observed for the GPT-3.5-turbo model. Indeed, the accuracy of this model increases significantly after receiving a hint, even if its initial accuracy was low. In contrast, the other models show a more moderate increase. When GPT-4o is used as the teacher, the hints provided by the error-based prompt seem more effective in improving the student models’ performance than those from the BaselineTwo prompt. However, the opposite effect is observed with Llama-3-8B-instruct as the teacher. Comparing the two teacher models, the figures suggest that the overall performance is better with Llama-3-8B-instruct. However, although accuracy is higher with Llama-3-8B-instruct, it would be crucial to verify the quality and relevance of the generated hints.

For more detailed results by exercise, you can refer to section \ref{subsec:add_res_exer}. There, we show how the accuracy evolves for each exercise. However, exercise 3 (module 7) is particularly challenging for the models to solve. Only the GPT student model manages to correct itself after being given a hint, whether the teacher is GPT-4o or Llama-3-8B-instruct.

\section{Discussion and Limitations}
%\subsection{Discussion}

% % --- summary of---------

Our work addresses a gap in hint generation research within the field of mathematics education. We demonstrated that language models could identify their own errors when acting as students, with error patterns varying based on the temperature setting. Higher temperatures led to more diverse outputs but increased errors, while lower temperatures produced more deterministic.
This error detection was crucial for selecting effective prompts for generating a synthetic hint dataset. We found that prompts focused on error correction and the BaselineTwo prompt were most successful. This aligns with the known challenges language models face with calculations and reasoning tasks.
Interestingly, our results differ from a previous study \citep{renze_effect_2024} on Multiple-Choice Question Answering (MCQA), which found that temperature did not impact problem-solving abilities but affected text variability. In contrast, we observed no clear link between temperature and problem-solving in our non-MCQA tasks, where we used only zero-shot approaches.

GPT-3.5-turbo showed the most effective self-correction after receiving optimized hints, likely due to the prompts being tailored for base GPT models, leading to a significant correction gap compared to other student models. Mistral-7B-instruct-v3 and Llama-3-8B-instruct, however, already had high accuracy with the hints, making further improvement harder, though their additional corrections remain noteworthy.

Notably, the Llama-3-8B-instruct model outperformed GPT-4o in accuracy when using the best prompts, challenging the assumption that larger models like GPT-4o are always superior. Future work should include a qualitative analysis of the generated hints in relation to pedagogical criteria and their relevance, as well as explore the potential of fine-tuning smaller models, such as Llama-3-8B-instruct, for hint generation.

This study has, however, several limitations. First, we only used four exercises from different modules, which is not sufficient for a comprehensive analysis, even though each exercise was solved 40 times. Results may differ with other exercise variants within these modules. Additionally, the cost of the API limited the number of exercises we could analyze.

The prompts for hint generation were optimized for GPT models, not for other models, so more tailored prompts might produce better results for non-GPT models. We also limited our analysis to GPT-4o and GPT-3.5-turbo due to cost constraints. Error type classification was evaluated only with GPT-4o, without the involvement of human experts, though some human verification was done. Including expert evaluation would provide deeper insights. Finally, the lack of qualitative analysis of the generated hints is another limitation, as such an analysis could offer valuable context and improve the overall assessment of hint quality.

\acks{We thank the anonymous reviewers for their constructive feedback. We also thanks the entire team at EvidenceB for hosting this project.}

\bibliography{pmlr-sample}

\begin{thebibliography}{59}
\providecommand{\natexlab}[1]{#1}
\providecommand{\url}[1]{\texttt{#1}}
\expandafter\ifx\csname urlstyle\endcsname\relax
  \providecommand{\doi}[1]{doi: #1}\else
  \providecommand{\doi}{doi: \begingroup \urlstyle{rm}\Url}\fi

\bibitem[cha()]{chatgpt_introducing_nodate}
Introducing {ChatGPT}.
\newblock URL \url{https://openai.com/index/chatgpt/}.

\bibitem[edu()]{education_recognizing_nodate}
Recognizing and {Responding} to {Cultural} {Differences} in the {Education} of {Culturally} and {Linguistically} {Diverse} {Learners} - {Steven} {P}. {Chamberlain}, 2005.
\newblock URL \url{https://journals.sagepub.com/doi/abs/10.1177/10534512050400040101}.

\bibitem[kha()]{khan_meet_nodate}
Meet {Khanmigo}: {Khan} {Academy}'s {AI}-powered teaching assistant \& tutor.
\newblock URL \url{https://www.khanacademy.org/khan-labs}.

\bibitem[qui()]{quizlet_q-chat_nodate}
Q-{Chat}: {Meet} your new {AI} tutor {\textbar} {Quizlet}.
\newblock URL \url{https://quizlet.com/qchat-personal-ai-tutor}.

\bibitem[Al-Hossami et~al.(2023)Al-Hossami, Bunescu, Teehan, Powell, Mahajan, and Dorodchi]{al-hossami_socratic_2023}
Erfan Al-Hossami, Razvan Bunescu, Ryan Teehan, Laurel Powell, Khyati Mahajan, and Mohsen Dorodchi.
\newblock Socratic {Questioning} of {Novice} {Debuggers}: {A} {Benchmark} {Dataset} and {Preliminary} {Evaluations}.
\newblock In Ekaterina Kochmar, Jill Burstein, Andrea Horbach, Ronja Laarmann-Quante, Nitin Madnani, Anaïs Tack, Victoria Yaneva, Zheng Yuan, and Torsten Zesch, editors, \emph{Proceedings of the 18th {Workshop} on {Innovative} {Use} of {NLP} for {Building} {Educational} {Applications} ({BEA} 2023)}, pages 709--726, Toronto, Canada, July 2023. Association for Computational Linguistics.
\newblock \doi{10.18653/v1/2023.bea-1.57}.
\newblock URL \url{https://aclanthology.org/2023.bea-1.57}.

\bibitem[Argyle et~al.(2023)Argyle, Busby, Fulda, Gubler, Rytting, and Wingate]{argyle_out_2023}
Lisa~P. Argyle, Ethan~C. Busby, Nancy Fulda, Joshua Gubler, Christopher Rytting, and David Wingate.
\newblock Out of {One}, {Many}: {Using} {Language} {Models} to {Simulate} {Human} {Samples}.
\newblock \emph{Political Analysis}, 31\penalty0 (3):\penalty0 337--351, July 2023.
\newblock ISSN 1047-1987, 1476-4989.
\newblock \doi{10.1017/pan.2023.2}.
\newblock URL \url{http://arxiv.org/abs/2209.06899}.
\newblock arXiv:2209.06899 [cs].

\bibitem[Arora et~al.(2022)Arora, Narayan, Chen, Orr, Guha, Bhatia, Chami, Sala, and Ré]{arora_ask_2022}
Simran Arora, Avanika Narayan, Mayee~F. Chen, Laurel Orr, Neel Guha, Kush Bhatia, Ines Chami, Frederic Sala, and Christopher Ré.
\newblock Ask {Me} {Anything}: {A} simple strategy for prompting language models, November 2022.
\newblock URL \url{http://arxiv.org/abs/2210.02441}.
\newblock arXiv:2210.02441 [cs].

\bibitem[Botelho et~al.(2023)Botelho, Baral, Erickson, Benachamardi, and Heffernan]{botelho_leveraging_2023}
Anthony Botelho, Sami Baral, John Erickson, Priyanka Benachamardi, and Neil Heffernan.
\newblock Leveraging natural language processing to support automated assessment and feedback for student open responses in mathematics.
\newblock \emph{Journal of Computer Assisted Learning}, 39:\penalty0 n/a--n/a, February 2023.
\newblock \doi{10.1111/jcal.12793}.

\bibitem[Bray(1999)]{bray_shadow_1999}
Mark Bray.
\newblock \emph{The {Shadow} education system: private tutoring and its implications for planners}.
\newblock January 1999.
\newblock Journal Abbreviation: http://lst-iiep.iiep-unesco.org/cgi-bin/wwwi32.exe/[in=epidoc1.in]/?t2000=026377/(100) Publication Title: http://lst-iiep.iiep-unesco.org/cgi-bin/wwwi32.exe/[in=epidoc1.in]/?t2000=026377/(100).

\bibitem[Brown et~al.(2020)Brown, Mann, Ryder, Subbiah, Kaplan, Dhariwal, Neelakantan, Shyam, Sastry, Askell, Agarwal, Herbert-Voss, Krueger, Henighan, Child, Ramesh, Ziegler, Wu, Winter, Hesse, Chen, Sigler, Litwin, Gray, Chess, Clark, Berner, McCandlish, Radford, Sutskever, and Amodei]{brown_language_2020}
Tom~B. Brown, Benjamin Mann, Nick Ryder, Melanie Subbiah, Jared Kaplan, Prafulla Dhariwal, Arvind Neelakantan, Pranav Shyam, Girish Sastry, Amanda Askell, Sandhini Agarwal, Ariel Herbert-Voss, Gretchen Krueger, Tom Henighan, Rewon Child, Aditya Ramesh, Daniel~M. Ziegler, Jeffrey Wu, Clemens Winter, Christopher Hesse, Mark Chen, Eric Sigler, Mateusz Litwin, Scott Gray, Benjamin Chess, Jack Clark, Christopher Berner, Sam McCandlish, Alec Radford, Ilya Sutskever, and Dario Amodei.
\newblock Language {Models} are {Few}-{Shot} {Learners}, July 2020.
\newblock URL \url{http://arxiv.org/abs/2005.14165}.
\newblock arXiv:2005.14165 [cs].

\bibitem[Bubeck et~al.(2023)Bubeck, Chandrasekaran, Eldan, Gehrke, Horvitz, Kamar, Lee, Lee, Li, Lundberg, Nori, Palangi, Ribeiro, and Zhang]{bubeck_sparks_2023}
Sébastien Bubeck, Varun Chandrasekaran, Ronen Eldan, Johannes Gehrke, Eric Horvitz, Ece Kamar, Peter Lee, Yin~Tat Lee, Yuanzhi Li, Scott Lundberg, Harsha Nori, Hamid Palangi, Marco~Tulio Ribeiro, and Yi~Zhang.
\newblock Sparks of {Artificial} {General} {Intelligence}: {Early} experiments with {GPT}-4, April 2023.
\newblock URL \url{http://arxiv.org/abs/2303.12712}.
\newblock arXiv:2303.12712 [cs].

\bibitem[Corbett and Anderson(1995)]{corbett_knowledge_1995}
Albert~T. Corbett and John~R. Anderson.
\newblock Knowledge tracing: {Modeling} the acquisition of procedural knowledge.
\newblock \emph{User Modelling and User-Adapted Interaction}, 4\penalty0 (4):\penalty0 253--278, 1995.
\newblock ISSN 0924-1868, 1573-1391.
\newblock \doi{10.1007/BF01099821}.
\newblock URL \url{http://link.springer.com/10.1007/BF01099821}.

\bibitem[Dzikovska et~al.(2010)Dzikovska, Moore, Steinhauser, Campbell, Farrow, and Callaway]{dzikovska_beetle_2010}
Myroslava~O. Dzikovska, Johanna~D. Moore, Natalie Steinhauser, Gwendolyn Campbell, Elaine Farrow, and Charles~B. Callaway.
\newblock Beetle {II}: {A} {System} for {Tutoring} and {Computational} {Linguistics} {Experimentation}.
\newblock In Sandra Kübler, editor, \emph{Proceedings of the {ACL} 2010 {System} {Demonstrations}}, pages 13--18, Uppsala, Sweden, July 2010. Association for Computational Linguistics.
\newblock URL \url{https://aclanthology.org/P10-4003}.

\bibitem[Dziri et~al.(2023)Dziri, Lu, Sclar, Li, Jiang, Lin, West, Bhagavatula, Bras, Hwang, Sanyal, Welleck, Ren, Ettinger, Harchaoui, and Choi]{dziri_faith_2023}
Nouha Dziri, Ximing Lu, Melanie Sclar, Xiang~Lorraine Li, Liwei Jiang, Bill~Yuchen Lin, Peter West, Chandra Bhagavatula, Ronan~Le Bras, Jena~D. Hwang, Soumya Sanyal, Sean Welleck, Xiang Ren, Allyson Ettinger, Zaid Harchaoui, and Yejin Choi.
\newblock Faith and {Fate}: {Limits} of {Transformers} on {Compositionality}, October 2023.
\newblock URL \url{http://arxiv.org/abs/2305.18654}.
\newblock arXiv:2305.18654 [cs].

\bibitem[Feng et~al.(2021)Feng, Magana, and Kao]{feng_systematic_2021}
Shi Feng, Alejandra Magana, and Dominic Kao.
\newblock A {Systematic} {Review} of {Literature} on the {Effectiveness} of {Intelligent} {Tutoring} {Systems} in {STEM}.
\newblock pages 1--9, October 2021.
\newblock \doi{10.1109/FIE49875.2021.9637240}.

\bibitem[Graesser et~al.(2004)Graesser, Lu, Jackson, Mitchell, Ventura, Olney, and Louwerse]{graesser_autotutor_2004}
Arthur Graesser, Shulan Lu, G.~Jackson, Heather Mitchell, Mathew Ventura, Andrew Olney, and Max Louwerse.
\newblock {AutoTutor}: a {Tutor} with {Dialogue} in {Natural} {Language}.
\newblock \emph{Behavior Research Methods}, 36:\penalty0 180--192, June 2004.
\newblock \doi{10.3758/BF03195563}.

\bibitem[Graesser et~al.(2012)Graesser, Conley, and Olney]{graesser_intelligent_2012}
Arthur Graesser, Mark Conley, and Andrew Olney.
\newblock Intelligent tutoring systems.
\newblock \emph{APA educational psychology handbook: Vol. 3. Applications to learning and teaching}, 3:\penalty0 451--473, January 2012.
\newblock ISSN 1-4338-0999-0.
\newblock \doi{10.1037/13275-018}.

\bibitem[Gros et~al.(2020)Gros, Thibaut, and Sander]{gros_semantic_2020}
Hippolyte Gros, Jean-Pierre Thibaut, and Emmanuel Sander.
\newblock Semantic congruence in arithmetic: {A} new conceptual model for word problem solving.
\newblock \emph{Educational Psychologist}, 55\penalty0 (2):\penalty0 69--87, April 2020.
\newblock ISSN 0046-1520, 1532-6985.
\newblock \doi{10.1080/00461520.2019.1691004}.
\newblock URL \url{https://www.tandfonline.com/doi/full/10.1080/00461520.2019.1691004}.

\bibitem[He-Yueya et~al.(2023)He-Yueya, Poesia, Wang, and Goodman]{he-yueya_solving_2023}
Joy He-Yueya, Gabriel Poesia, Rose~E. Wang, and Noah~D. Goodman.
\newblock Solving {Math} {Word} {Problems} by {Combining} {Language} {Models} {With} {Symbolic} {Solvers}, April 2023.
\newblock URL \url{http://arxiv.org/abs/2304.09102}.
\newblock arXiv:2304.09102 [cs].

\bibitem[Jangra et~al.(2024)Jangra, Mozafari, Jatowt, and Muresan]{jangra_navigating_2024}
Anubhav Jangra, Jamshid Mozafari, Adam Jatowt, and Smaranda Muresan.
\newblock Navigating the {Landscape} of {Hint} {Generation} {Research}: {From} the {Past} to the {Future}, April 2024.
\newblock URL \url{http://arxiv.org/abs/2404.04728}.
\newblock arXiv:2404.04728 [cs].

\bibitem[Jiang et~al.(2023)Jiang, Sablayrolles, Mensch, Bamford, Chaplot, de~las Casas, Bressand, Lengyel, Lample, Saulnier, Lavaud, Lachaux, Stock, Scao, Lavril, Wang, Lacroix, and Sayed]{jiang2023mistral7b}
Albert~Q. Jiang, Alexandre Sablayrolles, Arthur Mensch, Chris Bamford, Devendra~Singh Chaplot, Diego de~las Casas, Florian Bressand, Gianna Lengyel, Guillaume Lample, Lucile Saulnier, Lélio~Renard Lavaud, Marie-Anne Lachaux, Pierre Stock, Teven~Le Scao, Thibaut Lavril, Thomas Wang, Timothée Lacroix, and William~El Sayed.
\newblock Mistral 7b, 2023.
\newblock URL \url{https://arxiv.org/abs/2310.06825}.

\bibitem[Jiang et~al.(2021)Jiang, Toh, Molina, Donsbach, Cai, and Terry]{jiang_genline_2021}
Ellen Jiang, Edwin Toh, Alejandra Molina, Aaron Donsbach, Carrie Cai, and Michael Terry.
\newblock {GenLine} and {GenForm}: {Two} {Tools} for {Interacting} with {Generative} {Language} {Models} in a {Code} {Editor}.
\newblock pages 145--147, October 2021.
\newblock \doi{10.1145/3474349.3480209}.

\bibitem[Jiang et~al.(2022)Jiang, Olson, Toh, Molina, Donsbach, Terry, and Cai]{jiang_promptmaker_2022}
Ellen Jiang, Kristen Olson, Edwin Toh, Alejandra Molina, Aaron Donsbach, Michael Terry, and Carrie~J Cai.
\newblock {PromptMaker}: {Prompt}-based {Prototyping} with {Large} {Language} {Models}.
\newblock In \emph{{CHI} {Conference} on {Human} {Factors} in {Computing} {Systems} {Extended} {Abstracts}}, pages 1--8, New Orleans LA USA, April 2022. ACM.
\newblock ISBN 978-1-4503-9156-6.
\newblock \doi{10.1145/3491101.3503564}.
\newblock URL \url{https://dl.acm.org/doi/10.1145/3491101.3503564}.

\bibitem[Jin et~al.(2024)Jin, Liu, and Tan]{jin_zero-shot_2024}
Feihu Jin, Yifan Liu, and Ying Tan.
\newblock Zero-{Shot} {Chain}-of-{Thought} {Reasoning} {Guided} by {Evolutionary} {Algorithms} in {Large} {Language} {Models}, February 2024.
\newblock URL \url{http://arxiv.org/abs/2402.05376}.
\newblock arXiv:2402.05376 [cs].

\bibitem[Kinnebrew et~al.(2015)Kinnebrew, Segedy, and Biswas]{kinnebrew_integrating_2015}
John Kinnebrew, James Segedy, and Gautam Biswas.
\newblock Integrating {Model}-{Driven} and {Data}-{Driven} {Techniques} for {Analyzing} {Learning} {Behaviors} in {Open}-{Ended} {Learning} {Environments}.
\newblock \emph{IEEE Transactions on Learning Technologies}, 10:\penalty0 1--1, December 2015.
\newblock \doi{10.1109/TLT.2015.2513387}.

\bibitem[Knops(2022)]{knops2022neural}
Andr{\'e} Knops.
\newblock \emph{Neural and cognitive processes underlying numerosity perception and mental arithmetic}.
\newblock PhD thesis, Universit{\'e} Paris Cit{\'e}, 2022.

\bibitem[Kochmar et~al.(2020)Kochmar, Vu, Belfer, Gupta, Serban, and Pineau]{kochmar_automated_2020}
Ekaterina Kochmar, Dung~Do Vu, Robert Belfer, Varun Gupta, Iulian~Vlad Serban, and Joelle Pineau.
\newblock Automated {Personalized} {Feedback} {Improves} {Learning} {Gains} in an {Intelligent} {Tutoring} {System}, May 2020.
\newblock URL \url{http://arxiv.org/abs/2005.02431}.
\newblock arXiv:2005.02431 [cs].

\bibitem[Lan et~al.(2015)Lan, Vats, Waters, and Baraniuk]{lan_mathematical_2015}
Andrew~S. Lan, Divyanshu Vats, Andrew~E. Waters, and Richard~G. Baraniuk.
\newblock Mathematical {Language} {Processing}: {Automatic} {Grading} and {Feedback} for {Open} {Response} {Mathematical} {Questions}, January 2015.
\newblock URL \url{http://arxiv.org/abs/1501.04346}.
\newblock arXiv:1501.04346 [cs, stat].

\bibitem[Liu et~al.(2022)Liu, Wang, Baraniuk, and Lan]{liu_open-ended_2022}
Naiming Liu, Zichao Wang, Richard Baraniuk, and Andrew Lan.
\newblock Open-ended {Knowledge} {Tracing} for {Computer} {Science} {Education}.
\newblock In Yoav Goldberg, Zornitsa Kozareva, and Yue Zhang, editors, \emph{Proceedings of the 2022 {Conference} on {Empirical} {Methods} in {Natural} {Language} {Processing}}, pages 3849--3862, Abu Dhabi, United Arab Emirates, December 2022. Association for Computational Linguistics.
\newblock \doi{10.18653/v1/2022.emnlp-main.254}.
\newblock URL \url{https://aclanthology.org/2022.emnlp-main.254}.

\bibitem[Liu et~al.(2021)Liu, Yuan, Fu, Jiang, Hayashi, and Neubig]{liu_pre-train_2021}
Pengfei Liu, Weizhe Yuan, Jinlan Fu, Zhengbao Jiang, Hiroaki Hayashi, and Graham Neubig.
\newblock Pre-train, {Prompt}, and {Predict}: {A} {Systematic} {Survey} of {Prompting} {Methods} in {Natural} {Language} {Processing}, July 2021.
\newblock URL \url{http://arxiv.org/abs/2107.13586}.
\newblock arXiv:2107.13586 [cs].

\bibitem[Luo et~al.(2023)Luo, Sun, Xu, Zhao, Lou, Tao, Geng, Lin, Chen, and Zhang]{luo_wizardmath_2023}
Haipeng Luo, Qingfeng Sun, Can Xu, Pu~Zhao, Jianguang Lou, Chongyang Tao, Xiubo Geng, Qingwei Lin, Shifeng Chen, and Dongmei Zhang.
\newblock {WizardMath}: {Empowering} {Mathematical} {Reasoning} for {Large} {Language} {Models} via {Reinforced} {Evol}-{Instruct}, August 2023.
\newblock URL \url{http://arxiv.org/abs/2308.09583}.
\newblock arXiv:2308.09583 [cs].

\bibitem[Markel et~al.(2023)Markel, Opferman, Landay, and Piech]{markel_gpteach_2023}
Julia~M. Markel, Steven~G. Opferman, James~A. Landay, and Chris Piech.
\newblock {GPTeach}: {Interactive} {TA} {Training} with {GPT}-based {Students}.
\newblock In \emph{Proceedings of the {Tenth} {ACM} {Conference} on {Learning} @ {Scale}}, pages 226--236, Copenhagen Denmark, July 2023. ACM.
\newblock ISBN 9798400700255.
\newblock \doi{10.1145/3573051.3593393}.
\newblock URL \url{https://dl.acm.org/doi/10.1145/3573051.3593393}.

\bibitem[McNichols et~al.(2024)McNichols, Feng, Lee, Scarlatos, Smith, Woodhead, and Lan]{mcnichols_automated_2024}
Hunter McNichols, Wanyong Feng, Jaewook Lee, Alexander Scarlatos, Digory Smith, Simon Woodhead, and Andrew Lan.
\newblock Automated {Distractor} and {Feedback} {Generation} for {Math} {Multiple}-choice {Questions} via {In}-context {Learning}, January 2024.
\newblock URL \url{http://arxiv.org/abs/2308.03234}.
\newblock arXiv:2308.03234 [cs].

\bibitem[Mitrovic et~al.(2013)Mitrovic, Ohlsson, and Barrow]{mitrovic_effect_2013}
Antonija Mitrovic, Stellan Ohlsson, and Devon Barrow.
\newblock The effect of positive feedback in a constraint-based intelligent tutoring system.
\newblock \emph{Computers \& Education}, 60:\penalty0 264--272, January 2013.
\newblock \doi{10.1016/j.compedu.2012.07.002}.

\bibitem[Nguyen et~al.(2023)Nguyen, Stec, Hou, Di, and McLaren]{nguyen_evaluating_2023}
Huy~A. Nguyen, Hayden Stec, Xinying Hou, Sarah Di, and Bruce~M. McLaren.
\newblock Evaluating {ChatGPT}'s {Decimal} {Skills} and {Feedback} {Generation} in a {Digital} {Learning} {Game}, June 2023.
\newblock URL \url{http://arxiv.org/abs/2306.16639}.
\newblock arXiv:2306.16639 [cs].

\bibitem[Nie et~al.(2024)Nie, Chandak, Suzara, Ali, Woodrow, Peng, Sahami, Brunskill, and Piech]{nie_gpt_2024}
Allen Nie, Yash Chandak, Miroslav Suzara, Malika Ali, Juliette Woodrow, Matt Peng, Mehran Sahami, Emma Brunskill, and Chris Piech.
\newblock The {GPT} {Surprise}: {Offering} {Large} {Language} {Model} {Chat} in a {Massive} {Coding} {Class} {Reduced} {Engagement} but {Increased} {Adopters} {Exam} {Performances}, April 2024.
\newblock URL \url{http://arxiv.org/abs/2407.09975}.
\newblock arXiv:2407.09975 [cs, stat].

\bibitem[Park et~al.(2022)Park, Popowski, Cai, Morris, Liang, and Bernstein]{park_social_2022}
Joon~Sung Park, Lindsay Popowski, Carrie~J. Cai, Meredith~Ringel Morris, Percy Liang, and Michael~S. Bernstein.
\newblock Social {Simulacra}: {Creating} {Populated} {Prototypes} for {Social} {Computing} {Systems}, August 2022.
\newblock URL \url{http://arxiv.org/abs/2208.04024}.
\newblock arXiv:2208.04024 [cs].

\bibitem[Park et~al.(2023)Park, O'Brien, Cai, Morris, Liang, and Bernstein]{park_generative_2023}
Joon~Sung Park, Joseph~C. O'Brien, Carrie~J. Cai, Meredith~Ringel Morris, Percy Liang, and Michael~S. Bernstein.
\newblock Generative {Agents}: {Interactive} {Simulacra} of {Human} {Behavior}, August 2023.
\newblock URL \url{http://arxiv.org/abs/2304.03442}.
\newblock arXiv:2304.03442 [cs].

\bibitem[Phung et~al.(2023)Phung, Pădurean, Singh, Brooks, Cambronero, Gulwani, Singla, and Soares]{phung_automating_2023}
Tung Phung, Victor-Alexandru Pădurean, Anjali Singh, Christopher Brooks, José Cambronero, Sumit Gulwani, Adish Singla, and Gustavo Soares.
\newblock Automating {Human} {Tutor}-{Style} {Programming} {Feedback}: {Leveraging} {GPT}-4 {Tutor} {Model} for {Hint} {Generation} and {GPT}-3.5 {Student} {Model} for {Hint} {Validation}, December 2023.
\newblock URL \url{http://arxiv.org/abs/2310.03780}.
\newblock arXiv:2310.03780 [cs].

\bibitem[{Qinjin Jia} et~al.(2022){Qinjin Jia}, Young, {Yunkai Xiao}, {Jialin Cui}, {Chengyuan Liu}, Rashid, and Gehringer]{qinjin_jia_insta-reviewer_2022}
{Qinjin Jia}, Mitchell Young, {Yunkai Xiao}, {Jialin Cui}, {Chengyuan Liu}, Parvez Rashid, and Edward Gehringer.
\newblock Insta-{Reviewer}: {A} {Data}-{Driven} {Approach} for {Generating} {Instant} {Feedback} on {Students}' {Project} {Reports}.
\newblock July 2022.
\newblock \doi{10.5281/ZENODO.6853099}.
\newblock URL \url{https://zenodo.org/record/6853099}.
\newblock Publisher: Zenodo.

\bibitem[Razzaq et~al.(2020)Razzaq, Ostrow, and Heffernan]{razzaq_effect_2020}
Renah Razzaq, Korinn Ostrow, and Neil Heffernan.
\newblock Effect of {Immediate} {Feedback} on {Math} {Achievement} at the {High} {School} {Level}.
\newblock pages 263--267. June 2020.
\newblock ISBN 978-3-030-52239-1.
\newblock \doi{10.1007/978-3-030-52240-7_48}.

\bibitem[Renze and Guven(2024)]{renze_effect_2024}
Matthew Renze and Erhan Guven.
\newblock The {Effect} of {Sampling} {Temperature} on {Problem} {Solving} in {Large} {Language} {Models}, June 2024.
\newblock URL \url{http://arxiv.org/abs/2402.05201}.
\newblock arXiv:2402.05201 [cs].

\bibitem[Roest et~al.(2023)Roest, Keuning, and Jeuring]{roest_next-step_2023}
Lianne Roest, Hieke Keuning, and Johan Jeuring.
\newblock Next-{Step} {Hint} {Generation} for {Introductory} {Programming} {Using} {Large} {Language} {Models}, December 2023.
\newblock URL \url{http://arxiv.org/abs/2312.10055}.
\newblock arXiv:2312.10055 [cs].

\bibitem[Roller et~al.(2021)Roller, Dinan, Goyal, Ju, Williamson, Liu, Xu, Ott, Smith, Boureau, and Weston]{roller_recipes_2021}
Stephen Roller, Emily Dinan, Naman Goyal, Da~Ju, Mary Williamson, Yinhan Liu, Jing Xu, Myle Ott, Eric~Michael Smith, Y-Lan Boureau, and Jason Weston.
\newblock Recipes for {Building} an {Open}-{Domain} {Chatbot}.
\newblock In Paola Merlo, Jorg Tiedemann, and Reut Tsarfaty, editors, \emph{Proceedings of the 16th {Conference} of the {European} {Chapter} of the {Association} for {Computational} {Linguistics}: {Main} {Volume}}, pages 300--325, Online, April 2021. Association for Computational Linguistics.
\newblock \doi{10.18653/v1/2021.eacl-main.24}.
\newblock URL \url{https://aclanthology.org/2021.eacl-main.24}.

\bibitem[Ruan et~al.(2019)Ruan, Jiang, Xu, Tham, Qiu, Zhu, Murnane, Brunskill, and Landay]{ruan_quizbot_2019}
Sherry Ruan, Liwei Jiang, Justin Xu, Bryce Joe-Kun Tham, Zhengneng Qiu, Yeshuang Zhu, Elizabeth~L. Murnane, Emma Brunskill, and James~A. Landay.
\newblock {QuizBot}: {A} {Dialogue}-based {Adaptive} {Learning} {System} for {Factual} {Knowledge}.
\newblock In \emph{Proceedings of the 2019 {CHI} {Conference} on {Human} {Factors} in {Computing} {Systems}}, pages 1--13, Glasgow Scotland Uk, May 2019. ACM.
\newblock ISBN 978-1-4503-5970-2.
\newblock \doi{10.1145/3290605.3300587}.
\newblock URL \url{https://dl.acm.org/doi/10.1145/3290605.3300587}.

\bibitem[Sahoo et~al.(2024)Sahoo, Singh, Saha, Jain, Mondal, and Chadha]{sahoo_systematic_2024}
Pranab Sahoo, Ayush~Kumar Singh, Sriparna Saha, Vinija Jain, Samrat Mondal, and Aman Chadha.
\newblock A {Systematic} {Survey} of {Prompt} {Engineering} in {Large} {Language} {Models}: {Techniques} and {Applications}, February 2024.
\newblock URL \url{http://arxiv.org/abs/2402.07927}.
\newblock arXiv:2402.07927 [cs].

\bibitem[Scarlatos et~al.(2024)Scarlatos, Smith, Woodhead, and Lan]{scarlatos_improving_2024}
Alexander Scarlatos, Digory Smith, Simon Woodhead, and Andrew Lan.
\newblock Improving the {Validity} of {Automatically} {Generated} {Feedback} via {Reinforcement} {Learning}, March 2024.
\newblock URL \url{http://arxiv.org/abs/2403.01304}.
\newblock arXiv:2403.01304 [cs] version: 1.

\bibitem[Singh et~al.()Singh, Gulwani, and Solar-Lezama]{singh_automated_nodate}
Rishabh Singh, Sumit Gulwani, and Armando Solar-Lezama.
\newblock Automated {Feedback} {Generation} for {Introductory} {Programming} {Assignments}.

\bibitem[Song et~al.(2021)Song, Lee, and Oh]{song_context-aware_2021}
Dowon Song, Woosuk Lee, and Hakjoo Oh.
\newblock Context-aware and data-driven feedback generation for programming assignments.
\newblock pages 328--340, August 2021.
\newblock \doi{10.1145/3468264.3468598}.

\bibitem[Sonkar et~al.(2020)Sonkar, Waters, Lan, Grimaldi, and Baraniuk]{sonkar_qdkt_2020}
Shashank Sonkar, Andrew~E. Waters, Andrew~S. Lan, Phillip~J. Grimaldi, and Richard~G. Baraniuk.
\newblock {qDKT}: {Question}-centric {Deep} {Knowledge} {Tracing}, May 2020.
\newblock URL \url{http://arxiv.org/abs/2005.12442}.
\newblock arXiv:2005.12442 [cs, stat].

\bibitem[Sonkar et~al.(2023{\natexlab{a}})Sonkar, Le, Chen, Liu, Mallick, and Baraniuk]{sonkar_code_2023}
Shashank Sonkar, MyCo Le, Xinghe Chen, Naiming Liu, Debshila~Basu Mallick, and Richard~G. Baraniuk.
\newblock Code {Soliloquies} for {Accurate} {Calculations} in {Large} {Language} {Models}, October 2023{\natexlab{a}}.
\newblock URL \url{http://arxiv.org/abs/2309.12161}.
\newblock arXiv:2309.12161 [cs].

\bibitem[Sonkar et~al.(2023{\natexlab{b}})Sonkar, Liu, Mallick, and Baraniuk]{sonkar_class_2023}
Shashank Sonkar, Naiming Liu, Debshila~Basu Mallick, and Richard~G. Baraniuk.
\newblock {CLASS}: {A} {Design} {Framework} for building {Intelligent} {Tutoring} {Systems} based on {Learning} {Science} principles, October 2023{\natexlab{b}}.
\newblock URL \url{http://arxiv.org/abs/2305.13272}.
\newblock arXiv:2305.13272 [cs].

\bibitem[Steiss et~al.(2023)Steiss, {Tate}, Graham, {Cruz}, {Hebert}, Wang, Moon, {Tseng}, and {Warschauer}]{steiss_comparing_2023}
Jacob Steiss, {Tate}, Steve Graham, {Cruz}, {Hebert}, Jiali Wang, Youngsun Moon, {Tseng}, and {Warschauer}.
\newblock \emph{Comparing the {Quality} of {Human} and {ChatGPT} {Feedback} on {Students}' {Writing}}.
\newblock October 2023.
\newblock \doi{10.35542/osf.io/ty3em}.

\bibitem[SUAIB(2019)]{suaib_use_2019}
RAHMAH SUAIB.
\newblock {THE} {USE} {OF} {VISUAL} {AUDITORY} {KINESTHETIC} ({VAK}) {LEARNING} {STYLES} {TO} {INCREASE} {STUDENTS}’ {VOCABULARY}.
\newblock \emph{DIDAKTIKA}, 11:\penalty0 239, June 2019.
\newblock \doi{10.30863/didaktika.v11i2.169}.

\bibitem[Suraweera and Mitrovic(2002)]{suraweera_kermit_2002}
Pramuditha Suraweera and Antonija Mitrovic.
\newblock {KERMIT}: {A} constraint-based tutor for database modeling.
\newblock volume 2363, pages 377--387, June 2002.
\newblock ISBN 978-3-540-43750-5.
\newblock \doi{10.1007/3-540-47987-2_41}.

\bibitem[Touvron et~al.(2023{\natexlab{a}})Touvron, Lavril, Izacard, Martinet, Lachaux, Lacroix, Rozière, Goyal, Hambro, Azhar, Rodriguez, Joulin, Grave, and Lample]{touvron_llama_2023-1}
Hugo Touvron, Thibaut Lavril, Gautier Izacard, Xavier Martinet, Marie-Anne Lachaux, Timothée Lacroix, Baptiste Rozière, Naman Goyal, Eric Hambro, Faisal Azhar, Aurelien Rodriguez, Armand Joulin, Edouard Grave, and Guillaume Lample.
\newblock {LLaMA}: {Open} and {Efficient} {Foundation} {Language} {Models}, February 2023{\natexlab{a}}.
\newblock URL \url{http://arxiv.org/abs/2302.13971}.
\newblock arXiv:2302.13971 [cs].

\bibitem[Touvron et~al.(2023{\natexlab{b}})Touvron, Martin, Stone, Albert, Almahairi, Babaei, Bashlykov, Batra, Bhargava, Bhosale, Bikel, Blecher, Ferrer, Chen, Cucurull, Esiobu, Fernandes, Fu, Fu, Fuller, Gao, Goswami, Goyal, Hartshorn, Hosseini, Hou, Inan, Kardas, Kerkez, Khabsa, Kloumann, Korenev, Koura, Lachaux, Lavril, Lee, Liskovich, Lu, Mao, Martinet, Mihaylov, Mishra, Molybog, Nie, Poulton, Reizenstein, Rungta, Saladi, Schelten, Silva, Smith, Subramanian, Tan, Tang, Taylor, Williams, Kuan, Xu, Yan, Zarov, Zhang, Fan, Kambadur, Narang, Rodriguez, Stojnic, Edunov, and Scialom]{touvron_Llama_2023}
Hugo Touvron, Louis Martin, Kevin Stone, Peter Albert, Amjad Almahairi, Yasmine Babaei, Nikolay Bashlykov, Soumya Batra, Prajjwal Bhargava, Shruti Bhosale, Dan Bikel, Lukas Blecher, Cristian~Canton Ferrer, Moya Chen, Guillem Cucurull, David Esiobu, Jude Fernandes, Jeremy Fu, Wenyin Fu, Brian Fuller, Cynthia Gao, Vedanuj Goswami, Naman Goyal, Anthony Hartshorn, Saghar Hosseini, Rui Hou, Hakan Inan, Marcin Kardas, Viktor Kerkez, Madian Khabsa, Isabel Kloumann, Artem Korenev, Punit~Singh Koura, Marie-Anne Lachaux, Thibaut Lavril, Jenya Lee, Diana Liskovich, Yinghai Lu, Yuning Mao, Xavier Martinet, Todor Mihaylov, Pushkar Mishra, Igor Molybog, Yixin Nie, Andrew Poulton, Jeremy Reizenstein, Rashi Rungta, Kalyan Saladi, Alan Schelten, Ruan Silva, Eric~Michael Smith, Ranjan Subramanian, Xiaoqing~Ellen Tan, Binh Tang, Ross Taylor, Adina Williams, Jian~Xiang Kuan, Puxin Xu, Zheng Yan, Iliyan Zarov, Yuchen Zhang, Angela Fan, Melanie Kambadur, Sharan Narang, Aurelien Rodriguez, Robert Stojnic, Sergey Edunov, and Thomas
  Scialom.
\newblock Llama 2: {Open} {Foundation} and {Fine}-{Tuned} {Chat} {Models}, July 2023{\natexlab{b}}.
\newblock URL \url{http://arxiv.org/abs/2307.09288}.
\newblock arXiv:2307.09288 [cs].

\bibitem[Wang et~al.(2024)Wang, Zhang, Robinson, Loeb, and Demszky]{wang_bridging_2024}
Rose~E. Wang, Qingyang Zhang, Carly Robinson, Susanna Loeb, and Dorottya Demszky.
\newblock Bridging the {Novice}-{Expert} {Gap} via {Models} of {Decision}-{Making}: {A} {Case} {Study} on {Remediating} {Math} {Mistakes}, April 2024.
\newblock URL \url{http://arxiv.org/abs/2310.10648}.
\newblock arXiv:2310.10648 [cs].

\bibitem[Winkler and Söllner(2018)]{winkler_unleashing_2018}
Rainer Winkler and Matthias Söllner.
\newblock Unleashing the {Potential} of {Chatbots} in {Education}: {A} {State}-{Of}-{The}-{Art} {Analysis}.
\newblock \emph{Academy of Management Proceedings}, 2018:\penalty0 15903, April 2018.
\newblock \doi{10.5465/AMBPP.2018.15903abstract}.

\end{thebibliography}

\appendix
\newpage
\section{Additional information}
\label{ddrs}
%\addcontentsline{toc}{section}{Appendices}
\localtableofcontents
\newpage
%% write as you write sections
\subsection{Experimental pipelines}
This section outlines the pipelines used in our various experiments.

\begin{figure}[htbp]
    \centering
    \subfigure[Pipeline for determining the best prompt for hint generation]{
        \centering       \includegraphics[width=0.45\textwidth]{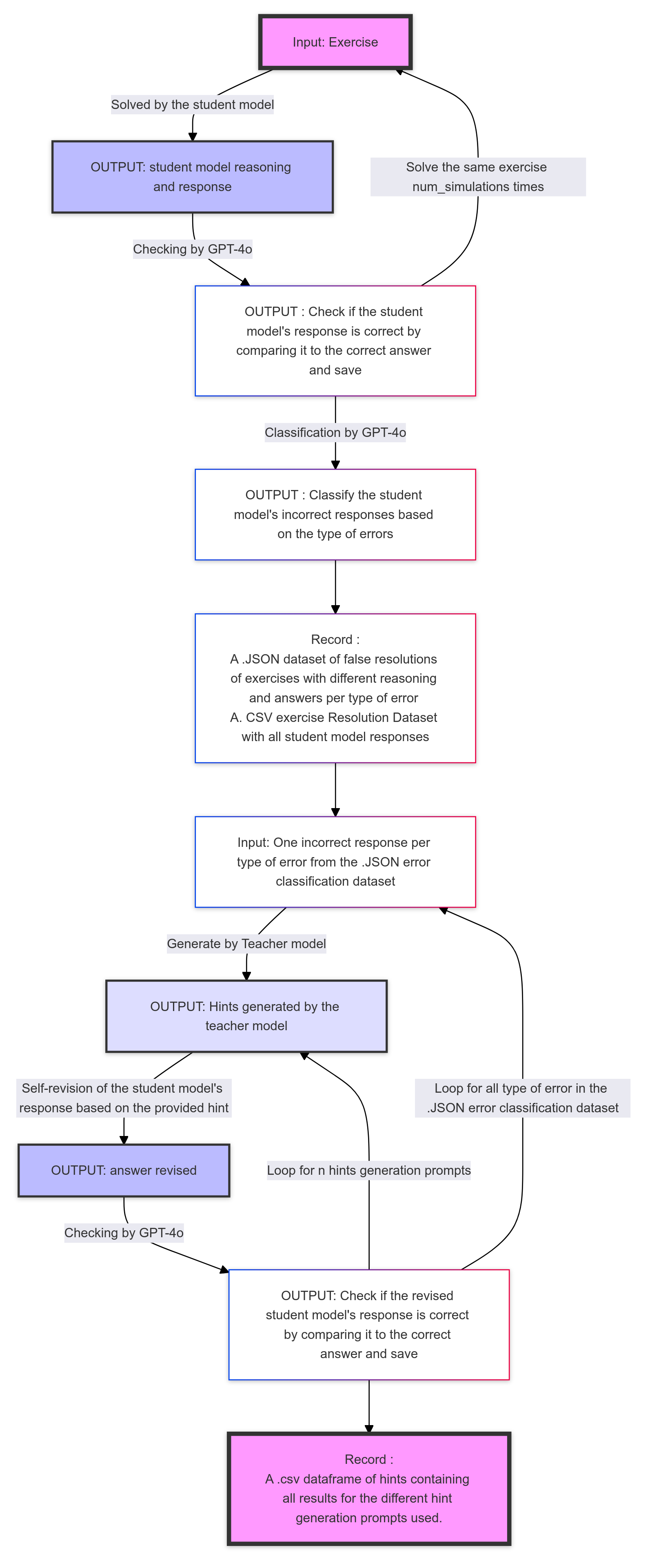}
        \label{fig:pipeline_select}
    }
    \hfill
    \subfigure[Pipeline for evaluating the best prompt]{
        \centering
   \includegraphics[width=0.45\textwidth]{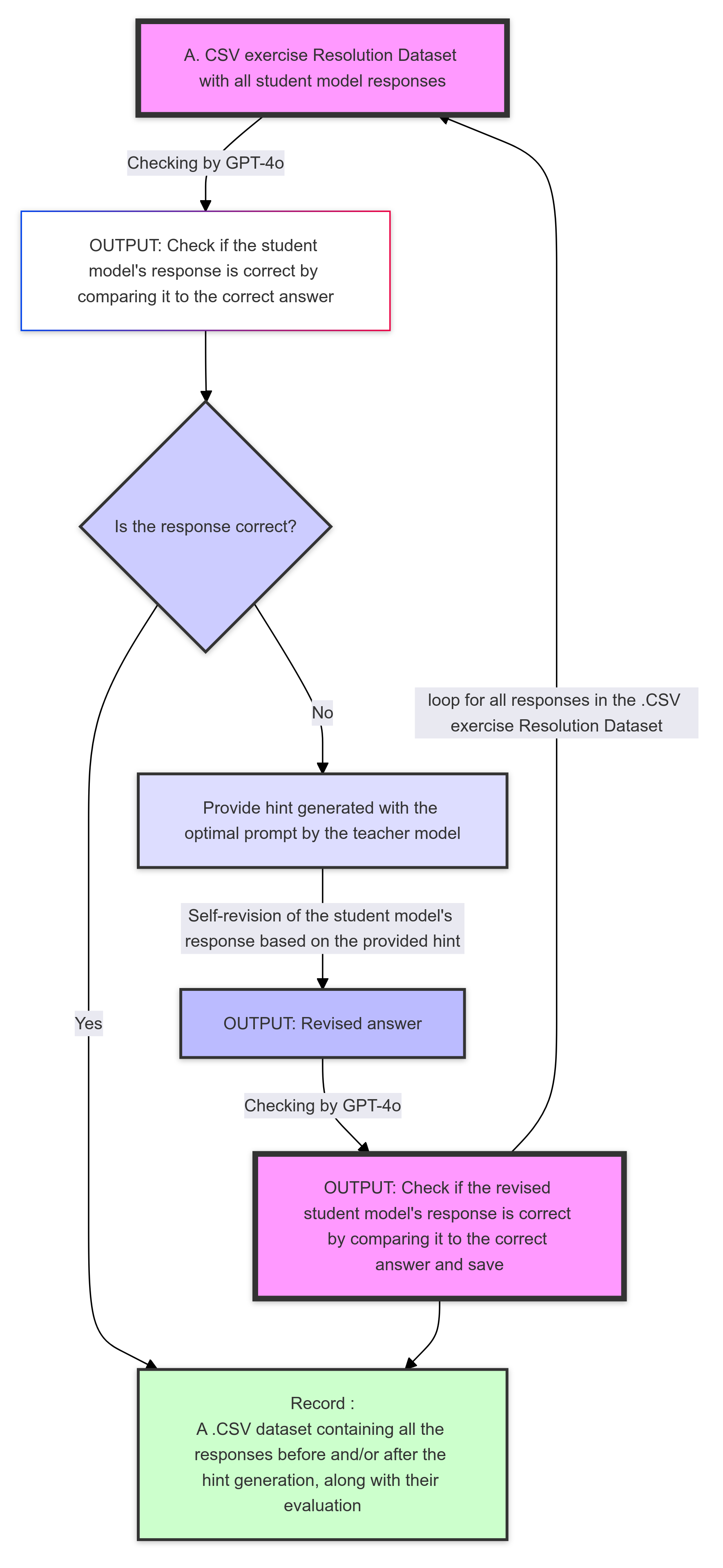}
        \label{fig:prompt_valid}
    }
\end{figure}

\subsection{Implementation details}
\label{subsec:implement}

We used the OpenAI API to interact with models based on GPT, such as GPT-4o and GPT-3.5-turbo. For other open-source models, like Llama-3-8B-Instruct and Mistral-7B-Instruct-v0.3, we accessed the resources via the HuggingFace platform. Specifically for these models, the prompts were executed using a 512 GB setup with two A100 GPUs.

During the exercise resolution phase and the review of the student's responses across various pipelines, several questions arise:

\textit{How can we ensure that the hints are given to the same student who made the error or provided the response? How can we guarantee that no previous hint is reused for the same student in the current session?}

The solution is to only add the student's initial solution to the context of the student for the current session. Indeed, whether using the API or open-source models, context management is done manually. There is no contextual dependency between requests unless we manually add the previous response to the context using the assistant role {“role”: “assistant”, “content”: “”}. By doing so, we ensure that the student who provided an incorrect response is the one who corrects their initial answer using the hints provided by the teacher.

\subsection{Prompts}
In this section, we present the various prompts used in the pipelines. These prompts were written in French in our experiments, but for the purposes of the paper, we have translated them into English.
\label{sec:prompts}
\subsubsection{Prompt for exercise resolution by the student model}
\label{sec:prompts_solve}

\begin{spverbatim}
 {"role": "system", "content": "You are a high school student who must solve mathematics exercises."},
{"role": "user", "content": '''Your objective is to answer the questions in the exercises by following the given instructions.
Exercise and question: {exercise}
Instructions: {instruct}
Required answer format: use a JSON format with the following structure:
{"reasoning": "Explain your reasoning here...", "answer": "Provide your answer here..."}
I emphasize that you must follow the required response format, and also that you must answer the questions in the exercises by following the instructions as given, without adding anything.'''}
\end{spverbatim}

\subsubsection{Prompt for answer review by the student model}
\label{sec:prompts_review}
\begin{spverbatim}
{"role": "system", "content": "You are a high school student who must solve mathematics exercises."}, 
{"role": "user", "content": '''You provided an incorrect answer to a math exercise. A teacher has given you a hint to help you understand your mistake and correct it. Your objective is to review your response to the questions in the exercise using the hint provided by the teacher.
Exercise and question: {exercise}.
Instruction: {instruct}
Hint: {hint}
Required response format: use a JSON format with the following structure: {{"response": "Write your answer here..."}}
Please provide a clean and readable output. I insist on this. Do not make formatting errors.
Respect the output format.'''}
\end{spverbatim}

\subsubsection{Prompt for classifying hints}
\label{sec:prompts_classif}
\begin{spverbatim}
{"role":"system", "content":"You are an expert in teaching mathematics"}, 
{"role": "user", "content":'''Your task is to verify if a student's revised answer to a mathematics exercise is correct or not by comparing it with the correct answer(s) provided. The exercises may have either a single correct answer or multiple correct answers.

The correct answer(s) for the exercise: {answer}
The student's revised answer: {revised_response} 
The hint: {hint} 
 1)- If the student's revised answer does not match the correct answer or any of the correct answers (if multiple), then put the hint in the "wrong_hint" field of the output.
 2)- If the student's revised answer includes at least one correct answer or all the correct answers, put the hint in the "correct_hint" field of the output. I insist on this, please follow this criterion.
 3)- If the hint contains the correct answer(s) or parts of the correct answer(s), then put the hint in the "wrong_hint" field of the output.
Put the output in a JSON format with the following structure: {{"correct_hint":"","wrong_hint":""}} 
Make sure that the generated output does not contain escape characters such as line breaks (\\n) or slashes (\\).
Please provide a clean and readable output. I insist on this. Do not make any formatting errors.
Follow the output format, and also follow the evaluation criteria and your role. Do not add anything else.'''}
\end{spverbatim}

\subsubsection{Prompt for checking if the answer is correct and detecting the type of error}
\label{sec:prompts_check}
\begin{spverbatim}
{"role": "system","content": "You are an expert in teaching mathematics"},
{"role": "user","content": '''Your task is to verify whether a student's answer to a mathematics exercise is correct or not by comparing it with the correct answer(s) provided. Exercises may have either a single correct answer or multiple correct answers.
The correct answer(s) for the exercise: {answer}
The student's answer: {student_answer}
The student's reasoning: {reasoning}
Categorize the student's error. Here are some categories of errors and examples. You can add other categories of errors. If the reasoning contains multiple errors, it is important to list all the present errors.
Specify each error distinctly, even if they belong to different categories or combine together.
1) Comprehension error: The student does not clearly understand the problem or the given instructions.
Example: Misreading a problem and confusing the given data.
2) Partial answer: The student provides part of the expected answer but fails to complete it correctly.
Example: In an equation with two variables, the student finds the value of one variable but forgets to find the value of the other.
3) Term grouping error: The student incorrectly combines or groups terms in a mathematical expression.
Example: When simplifying the expression 3x + 2x + 5, the student combines the terms 3x and 2x to get 5x^2 instead of 5x.
4) Simplification error: The student incorrectly simplifies a mathematical expression.
Example: When simplifying 6x/2, the student divides both the numerator and denominator by x instead of 2, resulting in an incorrect simplification of 6/2x.
5) Calculation error: The student incorrectly performs mathematical operations.
Example: When multiplying 7 by 8, the student gets 54 instead of 56.
6) Incorrect substitution error: The student substitutes an incorrect value into an expression or equation.
Example: In the equation 2x + 3y = 10, the student substitutes x = 4 instead of y = 2, leading to an incorrect solution.
7) Interpretation error: The student incorrectly interprets the problem’s instructions or data.
Example: In a probability problem, the student confuses the probability of event A with that of the complementary event of A.
8) Algebraic error: The student makes a mistake in algebraic manipulations, such as distributing, factoring, or solving equations.
Example: In solving 2(x + 3) = 10, the student incorrectly divides 10 by x + 3 instead of 2, leading to an incorrect answer.
1) - If the student's answer does not match the correct answer or any of the correct answers (if multiple), categorize the type of error and leave the "correct_answer" field empty.
2) - If the student's answer includes at least one correct answer or all the correct answers, put the student's answer in the "correct_answer" field. I insist on this, please follow this criterion.
Put the output in a JSON format with the following structure: {{"error_type": "","correct_answer": ""}}
Follow the output format, and also follow the evaluation criteria and your role. Do not add anything else.
    '''}
\end{spverbatim}

\subsubsection{Prompt for obtaining the diverse dataset with different reasoning and answers per type of errors}
\label{sec:prompts_diverse}
\begin{spverbatim}
{"role": "system", "content": "You are an expert in teaching mathematics"},
{"role": "user", "content": f'''
Your task is to classify a list of reasonings that contain multiple categories of errors. For each error category, you must provide the best examples with different reasoning.
In each reasoning, there may be multiple error categories. If that's the case, then you must find examples that are different for that group of error categories.

The error categories are already provided in the reasonings.
For each error category or group of error categories, you need to identify and provide the best k examples of different reasoning.
### Output Format
Make sure the generated output does not contain escape characters such as line breaks (\\n) or slashes (\\).
Please provide a clean and readable output. I insist on this. Do not make any formatting errors. Do not add errors that are not in the list.
You must provide the output in JSON format with the following structure:

        {{
            "different_reasoning": {{
                "category_1": [
                    {{"gpt_initial_reasoning": "","initial_response": "","evaluation": ""}},
                    ...
                ],
                "category_2": [
                    {{"gpt_initial_reasoning": "","initial_response": "","evaluation": ""}},
                    ...
                ],
                ...
            }}
        }}
The list of reasonings is: {list_reasoning}
Do not repeat the error groups, for example: calculation error, interpretation error is the same as interpretation error, calculation error.
The final JSON format must accurately reflect the classification you have made.
Please insert each reasoning into the appropriate category without modifying the content of the reasoning, the initial response, and the evaluation.'''}
\end{spverbatim}

\subsection{Prompt for hint generation}
\label{subsec:prompts_hint}
For the generation of hints, there are only a few differences in the user's role in each prompt. The rest of the content is identical, which is why we will include a complete example of one prompt. For the other prompts, we will only provide the user's role definition, specifying that the rest of the prompt follows the same structure.
\subsubsection{Baseline prompt}
\paragraph{BaselineOne prompt}
\label{par:completion_BOne}
\begin{spverbatim}
{"role": "system", "content": "You are an expert in teaching mathematics, helping students solve a math exercise by providing guiding hints following a specific cognitive approach."},
{"role": "user", "content": '''Your goal is to generate progressive hints to help students solve an exercise while following the specified cognitive approach. The hints should be given in increasing order of difficulty and should not reveal the final solution. The hints should encourage students to think independently while providing useful guidance. The hints must be in the form of questions, and they must not reveal the correct answer or any part of it—I insist on this.

The exercise and question: {exercise}.
Instruction: {instruct}
The correct answer to the exercise: {answer}
Guide according to the cognitive approach: {demarche_cog}
Required response format: use a JSON format with the following structure: {{"hints": ["hint1, hint2..."]}}. Do not number the hints.
I insist that you respect the response format and also ensure that the hints are in the form of questions and follow the specified cognitive approach. Provide only the hints, do not include any explanations.'''}
\end{spverbatim}

\paragraph{BaselineTwo prompt}
\label{par:completion_BTWO}
\begin{spverbatim}
"role": "user", "content": '''Your goal is to identify the common mistakes that students might make and to generate hints in the form of questions that can help them correct their mistakes and progress in solving the exercise. The hints must be in the form of questions, and they must not reveal the correct answer or any part of it, I insist on this...
\end{spverbatim}

\subsubsection{Prompt based on error type}  
\label{subsec:prompts_hint_err}
\paragraph{Prompt based on the student’s reasoning }
\label{par:completion_reason}
\begin{spverbatim}
{"role":"system", "content": "You are an expert in teaching mathematics"},
{"role": "user", "content":'''Your goal is to provide a clear and relevant hint to the student to help them correct their reasoning mistakes in math exercises. If the student has the correct answer, propose a hint to reinforce their understanding. This hint must be in the form of a question. Additionally, the hint must not include the correct answer to the exercise or any part of it.

The exercise and question: {exercise}.
Instruction: {instruct}
The correct answer to the exercise: {answer}
The student's reasoning: {gpt_reasoning}
The student's answer: {gpt_response}
Guide according to the cognitive approach: {demarche_cog}
JSON output format:  {{"hint": "Place the hint here without numbering it..."}} 
Ensure the generated output does not contain escape characters such as line breaks (\\n) or slashes (\\). 
Please provide a clean and readable output. I insist on this. Do not make any formatting errors.
Follow the output format. I emphasize that the hint should not be numbered and must be in the form of a question. '''}
\end{spverbatim}

\paragraph{Prompt based on the method used by the student }
\label{par:completion_meth}
\begin{spverbatim}
"role": "user", "content":'''Your goal is to provide a hint that helps the student review the method they are using to solve the math exercise. If the student has a correct method, propose a hint to reinforce their understanding of that method. This hint must be in the form of a question. Additionally, the hint must not include the correct answer to the exercise or any part of it...
\end{spverbatim}

\paragraph{Prompt based on the application of concepts}
\label{par:completion_concept}
\begin{spverbatim}
"role": "user", "content":'''Your goal is to provide a hint that helps the student review the application of mathematical concepts to solve the exercise and find the correct answer. If the student is applying the concepts correctly, propose a hint to reinforce their understanding. This hint must be in the form of a question. Additionally, the hint must not include the correct answer to the exercise or any part of it...
\end{spverbatim}

 \paragraph{Prompt based on calculations}
\label{par:completion_calcul}
\begin{spverbatim}
"role": "user", "content":'''Your goal is to provide a hint that helps the student review the calculations performed to solve the math exercise. If the student's calculations are correct, propose a hint to reinforce their understanding of the calculation steps. This hint must be in the form of a question. Additionally, the hint must not include the correct answer to the exercise or any part of it...
\end{spverbatim}

\paragraph{Prompt based on problem interpretation}
\label{par:completion_interp}
\begin{spverbatim}
"role": "user", "content":'''Your goal is to provide a hint that helps the student review their interpretation of the math problem. If the student interprets the problem correctly, propose a hint to reinforce their understanding. This hint must be in the form of a question. Additionally, the hint must not include the correct answer to the exercise or any part of it...
\end{spverbatim}

\paragraph{Prompt combining all the above aspects }
\label{par:completion_all}
\begin{spverbatim}
"role": "user", "content":'''Your goal is to provide a clear and relevant hint to the student to help them correct their mistakes and improve their answers in math exercises. This hint must be in the form of a question. Additionally, the hint must not include the correct answer to the exercise or any part of it.
Consider the following aspects when generating the hint:
   - Reasoning
   - Method
   - Application of concepts
   - Calculations
   - Interpretation of the problem ...
\end{spverbatim}

\paragraph{Prompt for completing partial answers}
\label{par:completion_part}
\begin{spverbatim}
"role": "user", "content":'''Your goal is to provide a clear and relevant hint to the student to help them complete their partial answer in math exercises. This hint must be in the form of a question. Additionally, the hint must not include the correct answer to the exercise or any part of it...
\end{spverbatim}

\subsection{Metrics and Additional Results}
\label{sec:add_res}

Metrics such as accuracy, error rate, revision success rate, and mean revision error  were used to quantify the performance of the models in order to determine the best prompts.

\begin{itemize}
    \item Accuracy: For each temperature, we calculate the accuracy as the number of correct responses out of the 40 repetitions divided by the number of responses (number of correct+number of incorrect responses). $$
\text{Accuracy} = \frac{\text{Number of Correct Responses}}{\text{number of responses}}
$$
    \item Revision Success Rate: For a specific type of error at a specific temperature, this is calculated as the number of correct responses out of the 10 repetitions divided by the number of responses. $$
\text{Revision Success Rate} = \frac{\text{Number of Correct Responses}}{\text{number of responses}}
$$

    \item Error Revsion Rate: The revision error rate is calculated as: $$
\text{Error Revision Rate} = 1 - \text{Revision Success Rate}
$$

    \item Mean Revision Error Rate: For all encountered errors at a specific temperature, we calculate the mean revision error rate as the sum of individual error rates divided by the number of errors encountered. $$
\text{Mean Revision Error Rate} = \frac{\sum_{i=1}^{N} \text{ErrorRevisionRate}_i}{N}
$$
  where \( N \) is the number of different types of errors encountered.
\end{itemize}

\subsubsection{Additional results across exercises based on temperature for student model accuracy before and after hints guided by Llama-3-8B-Instruct vs. GPT-4o}
\label{subsec:add_res_exer}
\begin{figure}[htbp]
    \centering
     \begin{minipage}[b]{0.6\textwidth}
        \centering
        \includegraphics[width=\textwidth]{images/experiments_prompt_validation/Screenshot_2024-09-06_at_13.05.12.png}
    \end{minipage}
    \vspace{1em}
    \subfigure[GPT-4o]{
        \includegraphics[width=0.5\textwidth]{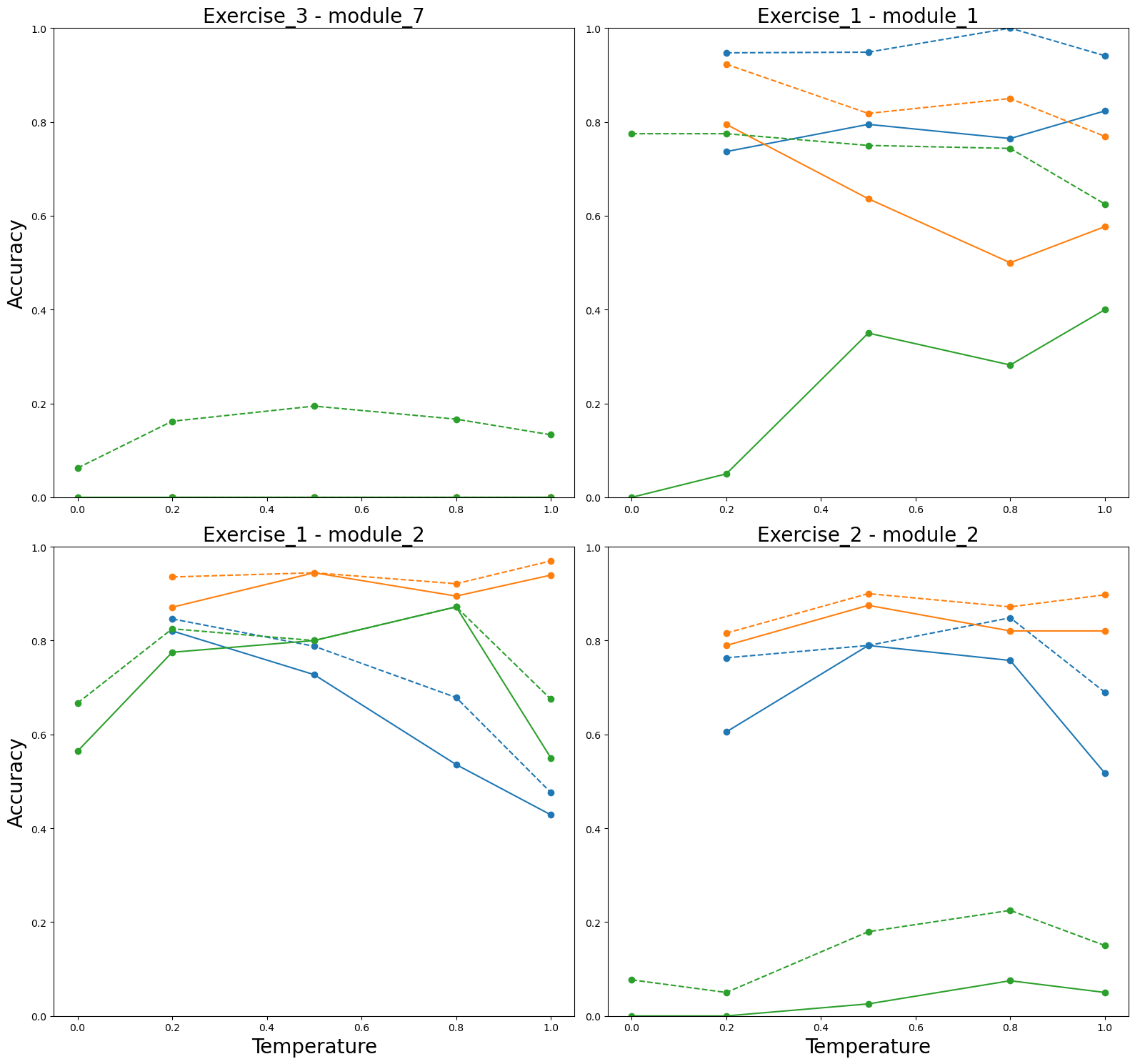}
    }
    \hfill
    \subfigure[Llama-3-8B-instruct]{
        \includegraphics[width=0.5\textwidth]{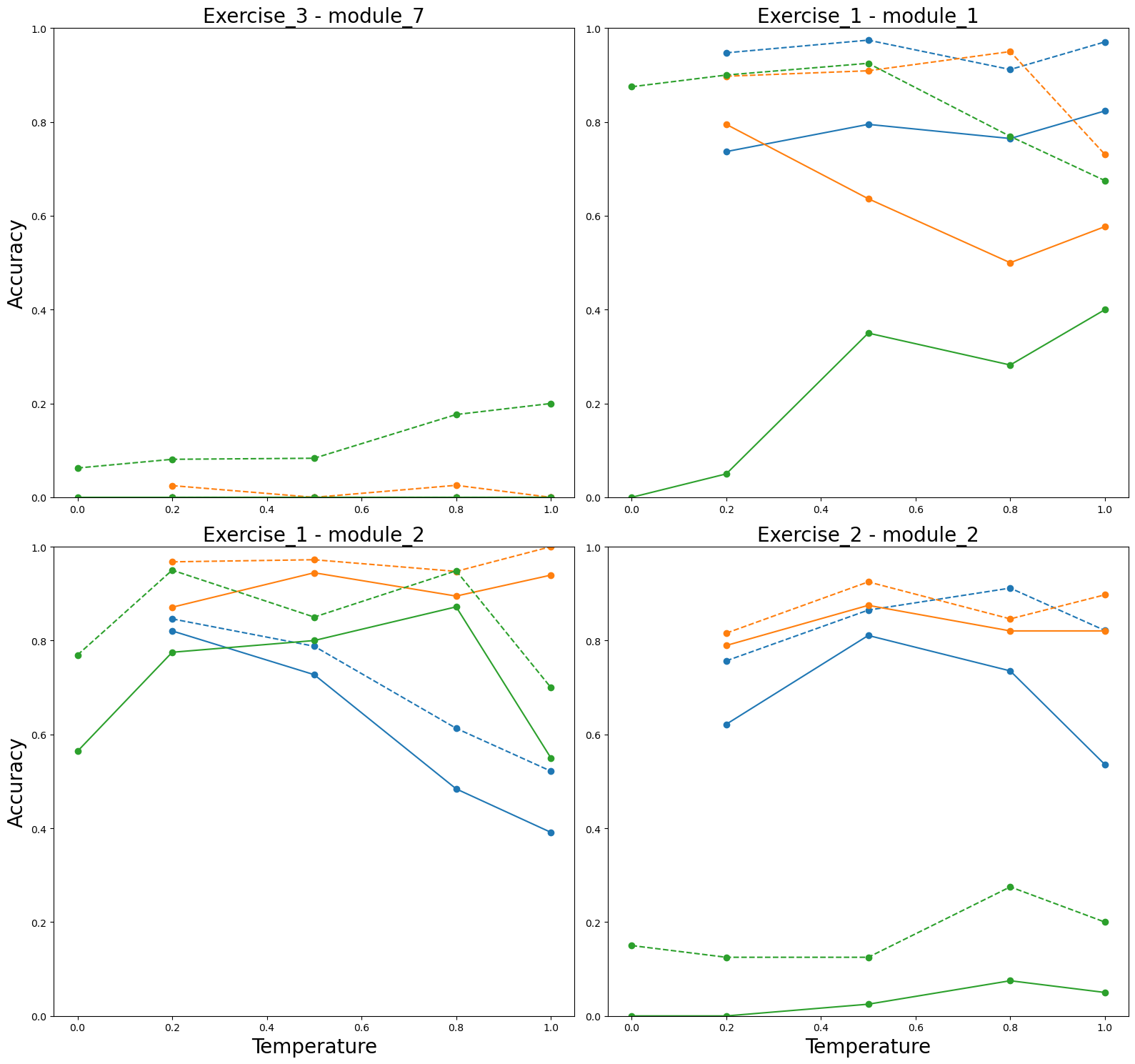}
    }
    \caption{Comparison of accuracy before and after providing hints across each exercise for different student models, using GPT-4o and Llama-3-8B-instruct as teacher models with \textbf{the best specialized hint generation prompt focused on calculation errors}. The results show improved performance when using Llama-3-8B-instruct as the teacher model, but the student models struggled to correct themselves on exercise 3 - module 7.}
    \label{fig:comp_best_specialized_prompt}
\end{figure}

\begin{figure}[htbp]
     \centering
    \begin{minipage}[b]{0.6\textwidth}
        \centering
        \includegraphics[width=\textwidth]{images/experiments_prompt_validation/Screenshot_2024-09-06_at_13.05.12.png}
    \end{minipage}
    \vspace{1em}
    \subfigure[GPT-4o]{
        \includegraphics[width=0.5\textwidth]{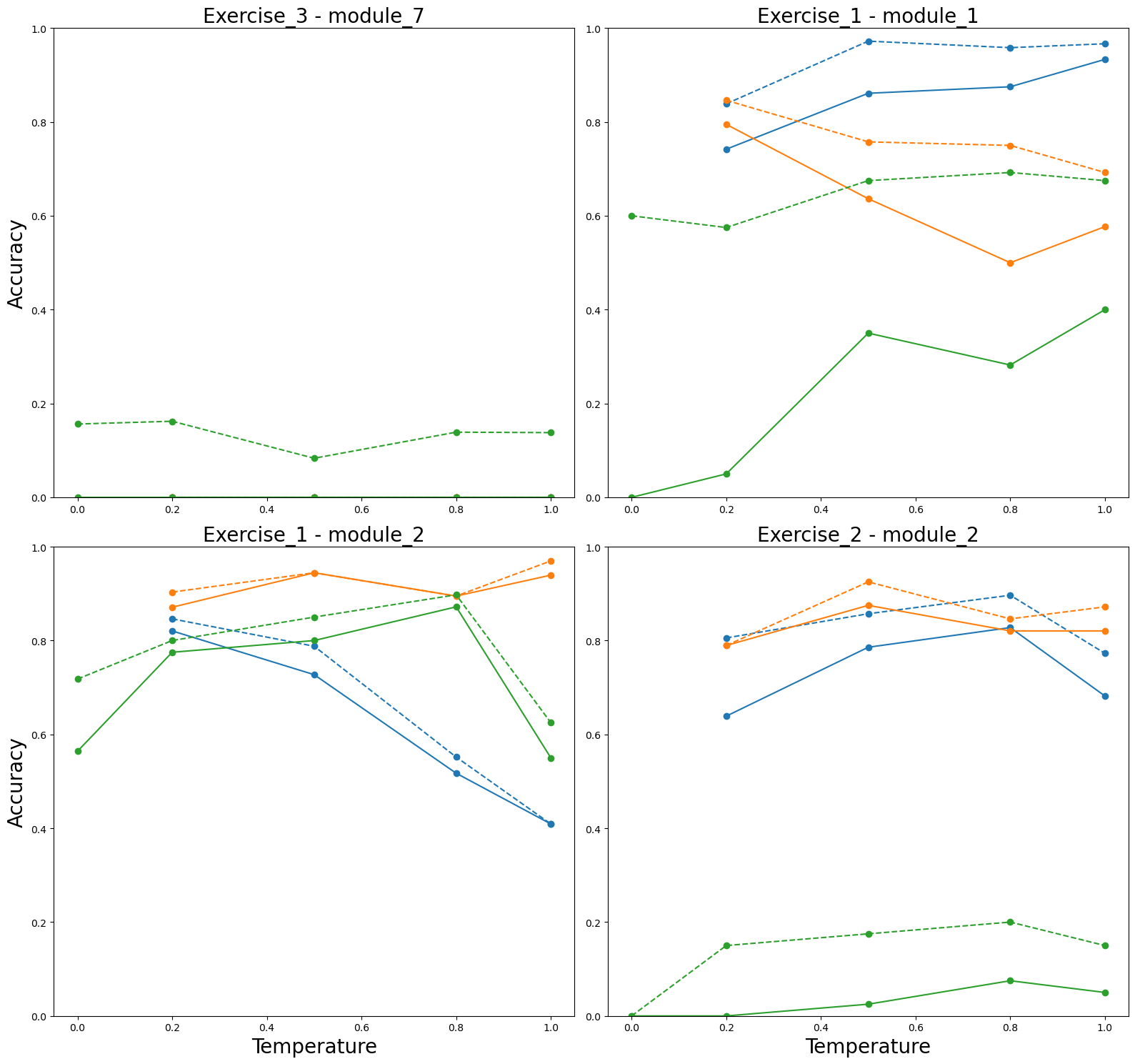}
    }
    \hfill
    \subfigure[Llama-3-8B-instruct]{
        \includegraphics[width=0.5\textwidth]{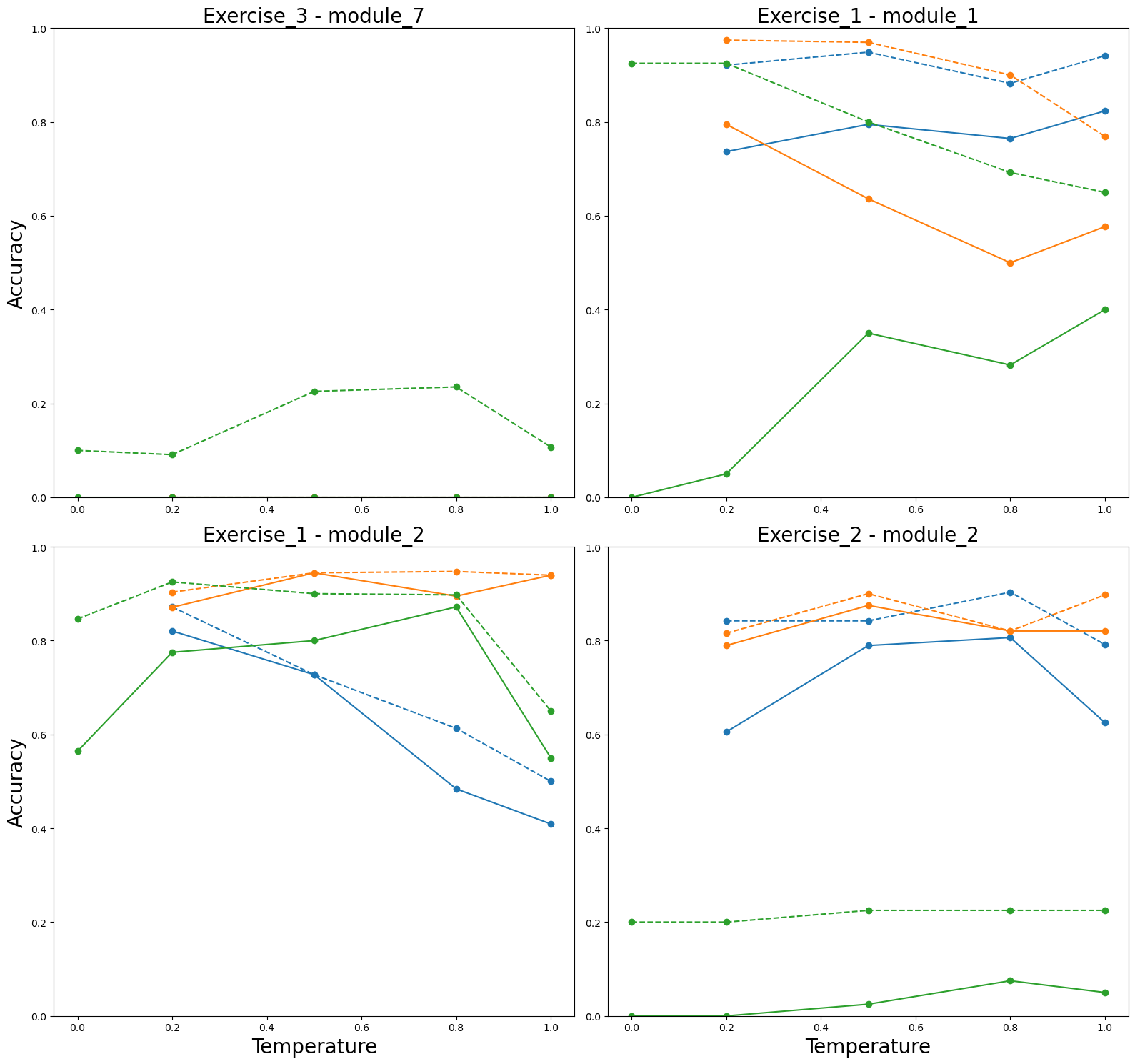}
    }
    \caption{Comparison of accuracy before and after providing hints across each exercise for different student models, using GPT-4o and Llama-3-8B-instruct as teacher models with \textbf{the best baseline-type hint generation prompt, named BaselineTwo.} The results show improved performance when using Llama-3-8B-instruct as the teacher model, but the student models struggled to correct themselves on exercise 3 - module 7. }
    \label{fig:comp_best_open_prompt}
\end{figure}

Comparison of accuracy before and after providing hints across each exercise for different student models using different teachers models with the best specialized and best baseline-type hint generation prompts is shown in Figure \ref{fig:comp_best_open_prompt} and \ref{fig:comp_best_specialized_prompt}.

\subsubsection{Additional results on mean revision error rate by temperature and prompt for each student model across other exercises}
\label{sec:add_res_mean}
\begin{figure}[htbp]
    \centering
    \subfigure[GPT-3.5-turbo]{
        \includegraphics[width=0.65\textwidth]{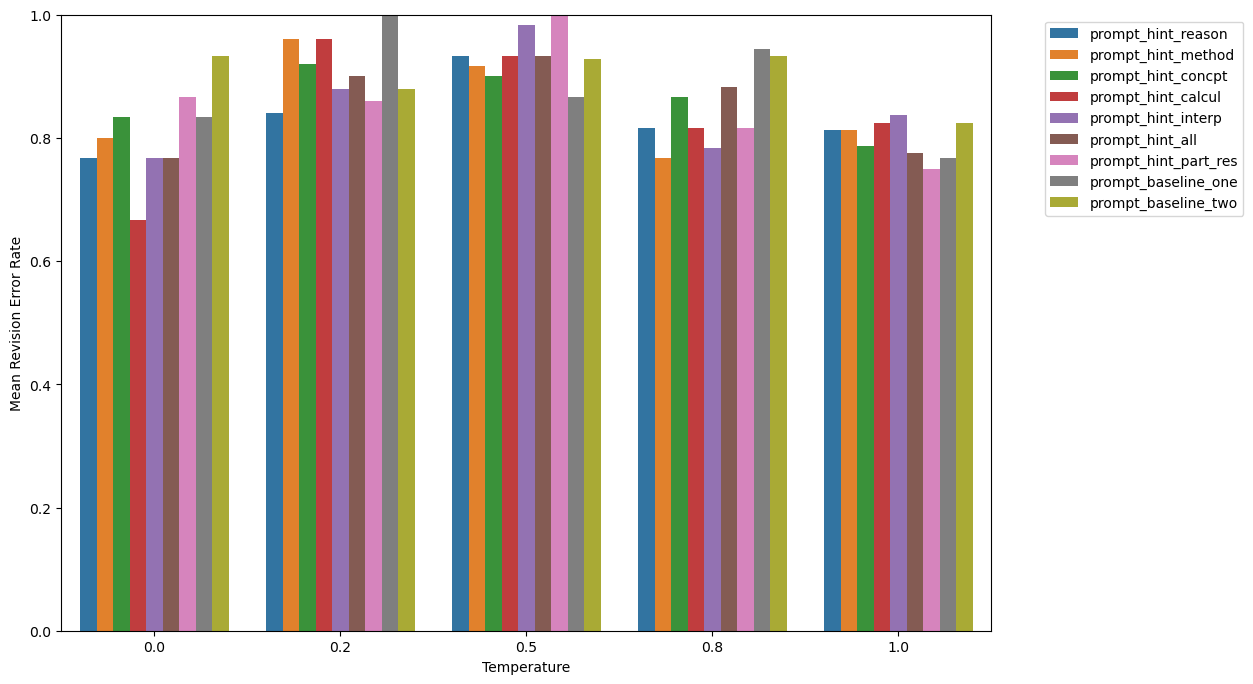}
    }
    %\hfill
    \subfigure[Llama-3-8B-instruct]{
        \includegraphics[width=0.65\textwidth]{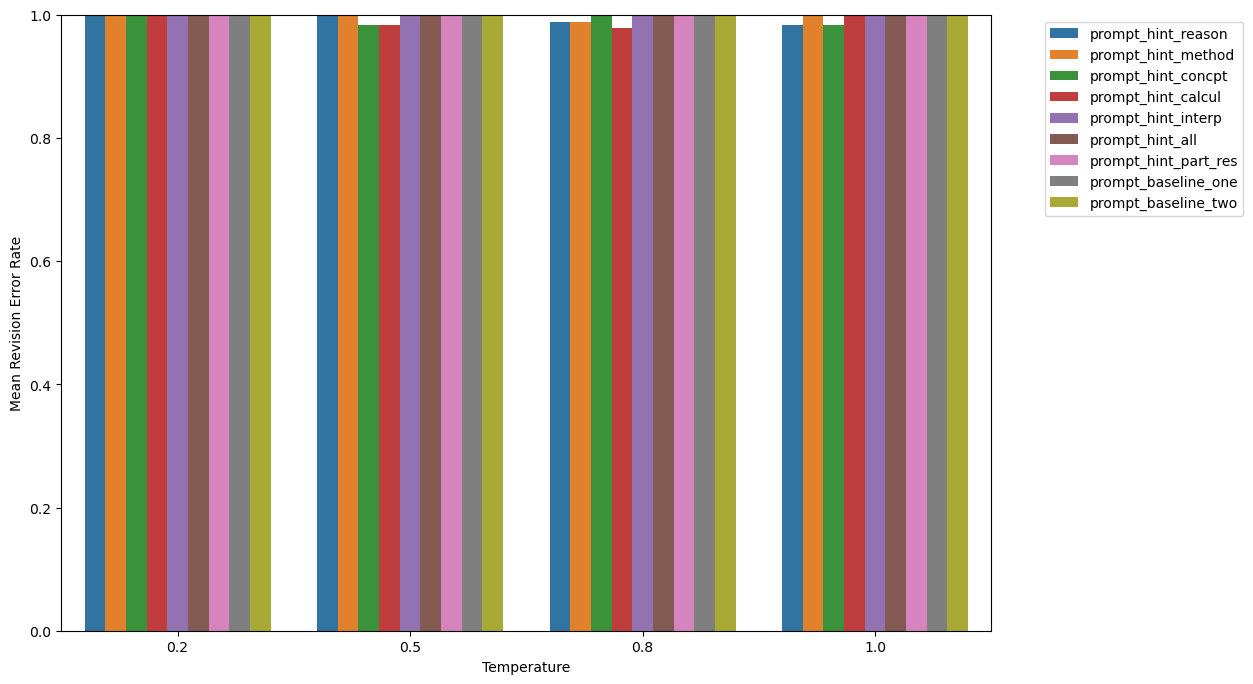}
    }
    %\hfill
    \subfigure[Mistral-7B-Instruct-v0.3]{
        \includegraphics[width=0.65\textwidth]{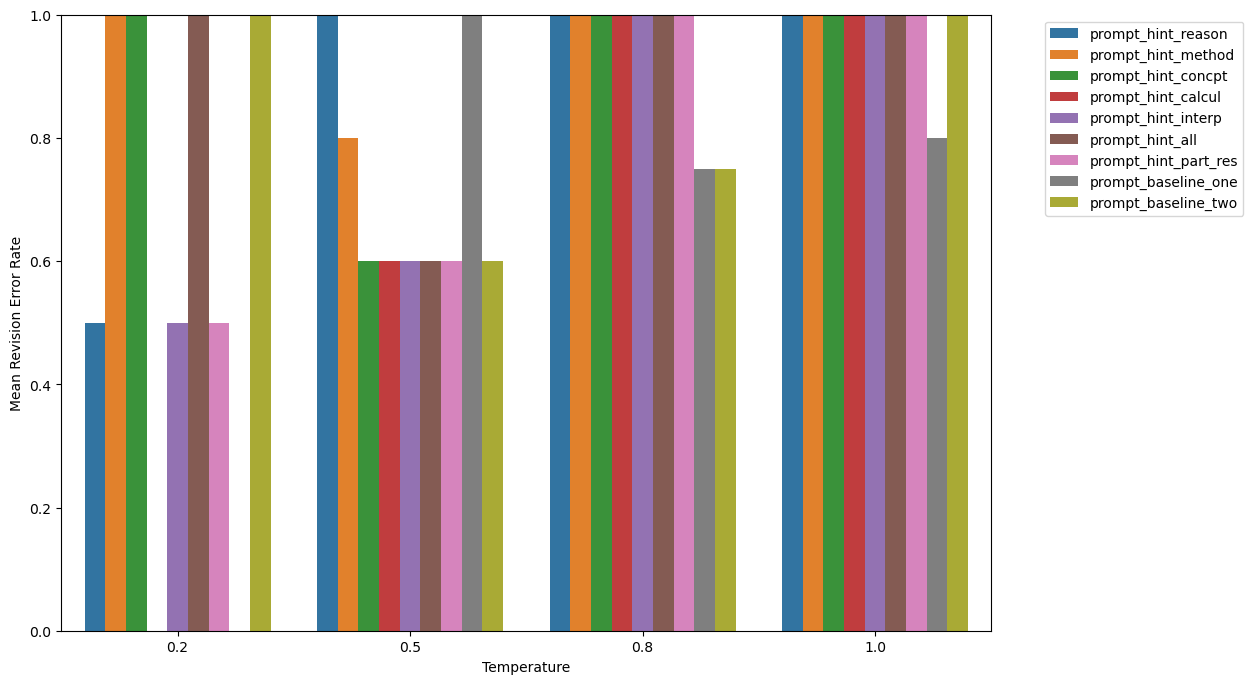}
    }
    \caption{Comparison of mean revision error rates across different temperatures and prompts for hint generation in three student models (GPT-3.5-turbo, Llama-3-8B-instruct, Mistral-7B-instruct-v3) on exercise 3 - module 7. This Figures hows that this particular exercise is challenging to solve, even with hints, for all student models. The accuracies before revision are 0.00 for all temperatures and models: \textbf{GPT-3.5-turbo}, \textbf{Llama-3-8B-instruct}, and \textbf{Mistral-7B-instruct-v3}}
    \label{fig:mean_error_ex3mod7}  
\end{figure}

\begin{figure}[htbp]
    \centering
    \subfigure[GPT-3.5-turbo]{
        \includegraphics[width=0.65\textwidth]{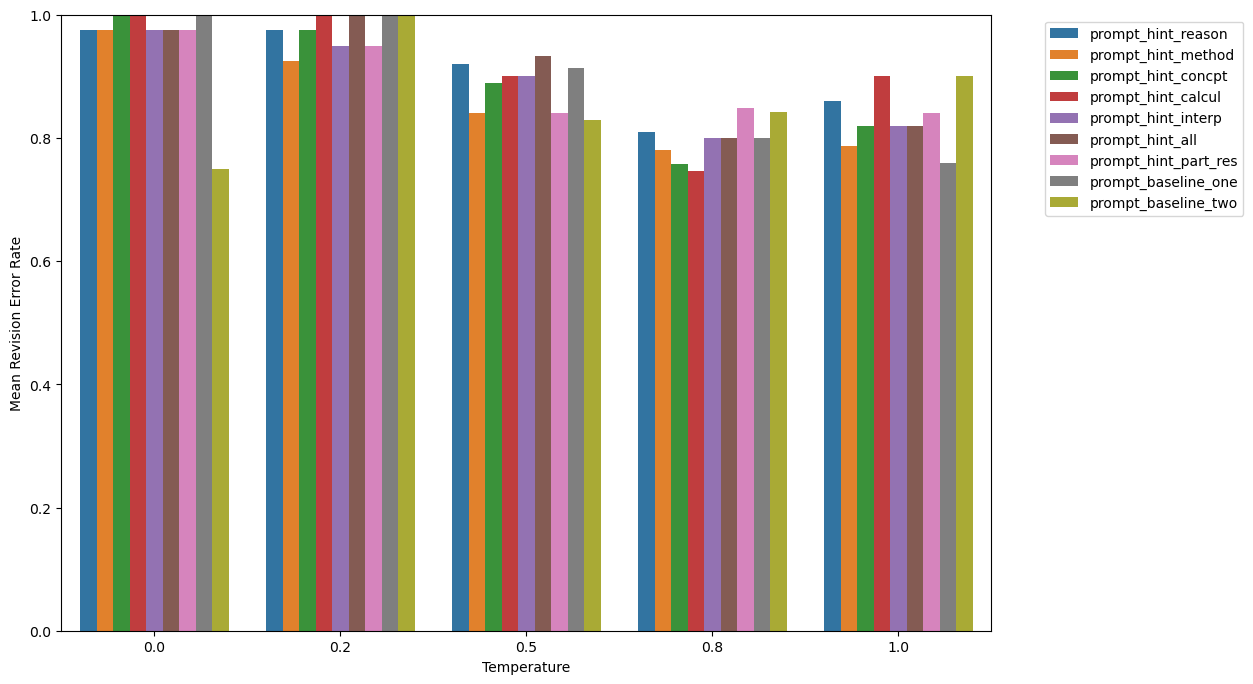}
    }
    %\hfill
    \subfigure[Llama-3-8B-instruct]{
        \includegraphics[width=0.65\textwidth]{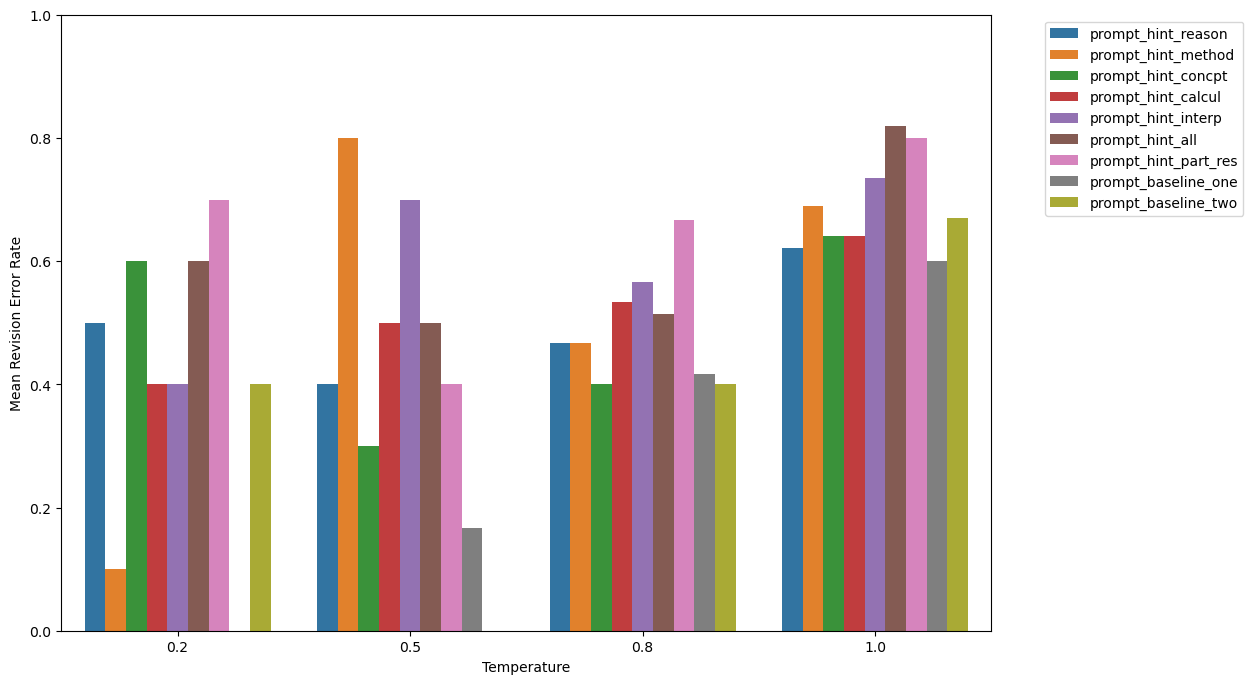}
    }
    %\hfill
    \subfigure[Mistral-7B-Instruct-v0.3]{
        \includegraphics[width=0.65\textwidth]{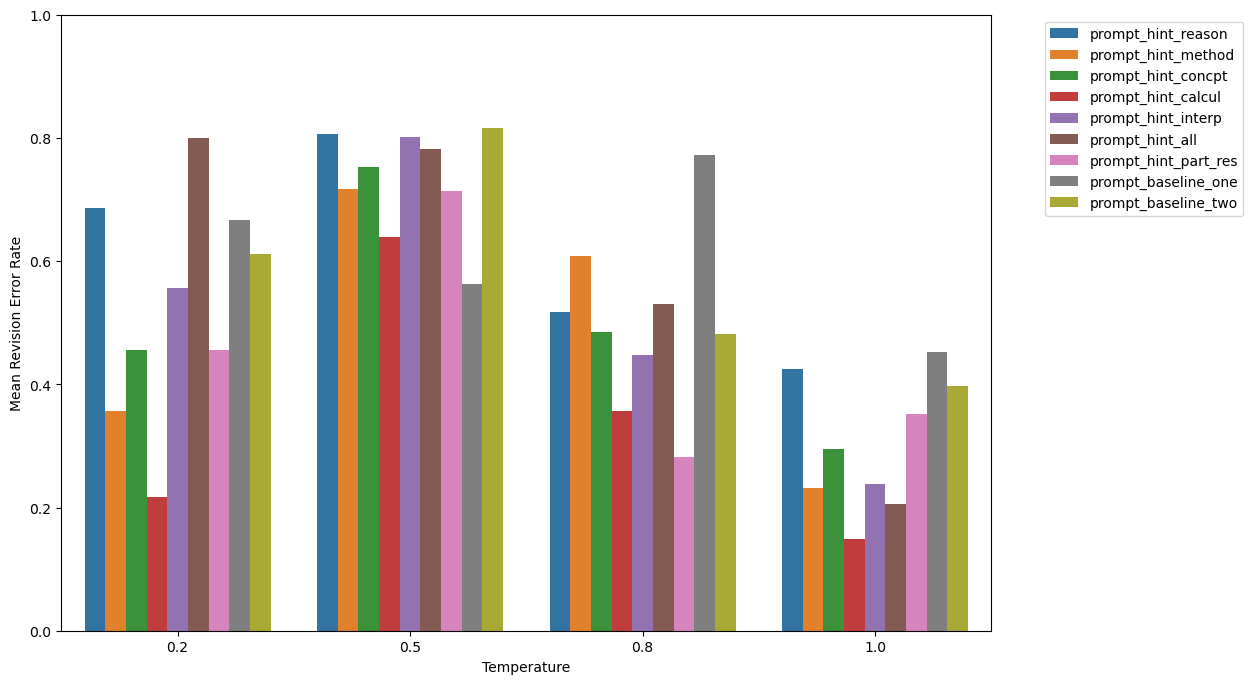}
    }
    \caption{Comparison of mean revision error rates across different temperatures and prompts for hint generation in three student models (GPT-3.5-turbo, Llama-3-8B-instruct, Mistral-7B-instruct-v3) on exercise 2 - module 2. The Figure illustrates how each model’s mean revision error rate evolves as the temperature increases and with varying prompts, highlighting how effectively the models can correct themselves using the provided hints. The accuracies before revision are as follows: For \textbf{GPT-3.5-turbo}, the accuracies are 0.00 at a temperature of 0.0, 0.00 at 0.2, 0.025 at 0.5, 0.075 at 0.8, and 0.05 at 1.0. For \textbf{Llama-3-8B-instruct}, the accuracies are 0.816 at 0.2, 0.875 at 0.5, 0.821 at 0.8, and 0.9 at 1.0. \textbf{Mistral-7B-instruct-v3} exhibits accuracies of 0.605 at 0.2, 0.789 at 0.5, 0.714 at 0.8, and 0.517 at 1.0.}
    \label{fig:mean_error_ex2mod2}
\end{figure}

\begin{figure}[htbp]
    \centering
    \subfigure[GPT-3.5-turbo]{
        \includegraphics[width=0.65\textwidth]{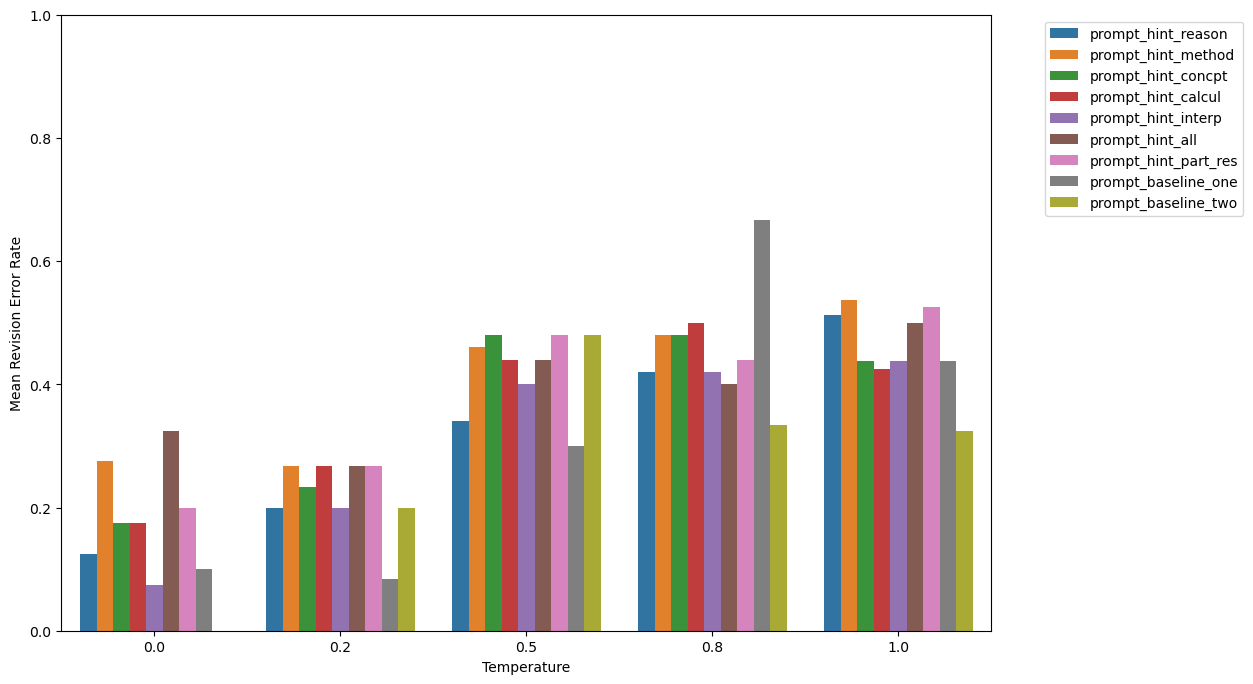}
        \label{fig:gpt}
    }
    \hfill
    \subfigure[Llama-3-8B-instruct]{
        \includegraphics[width=0.65\textwidth]{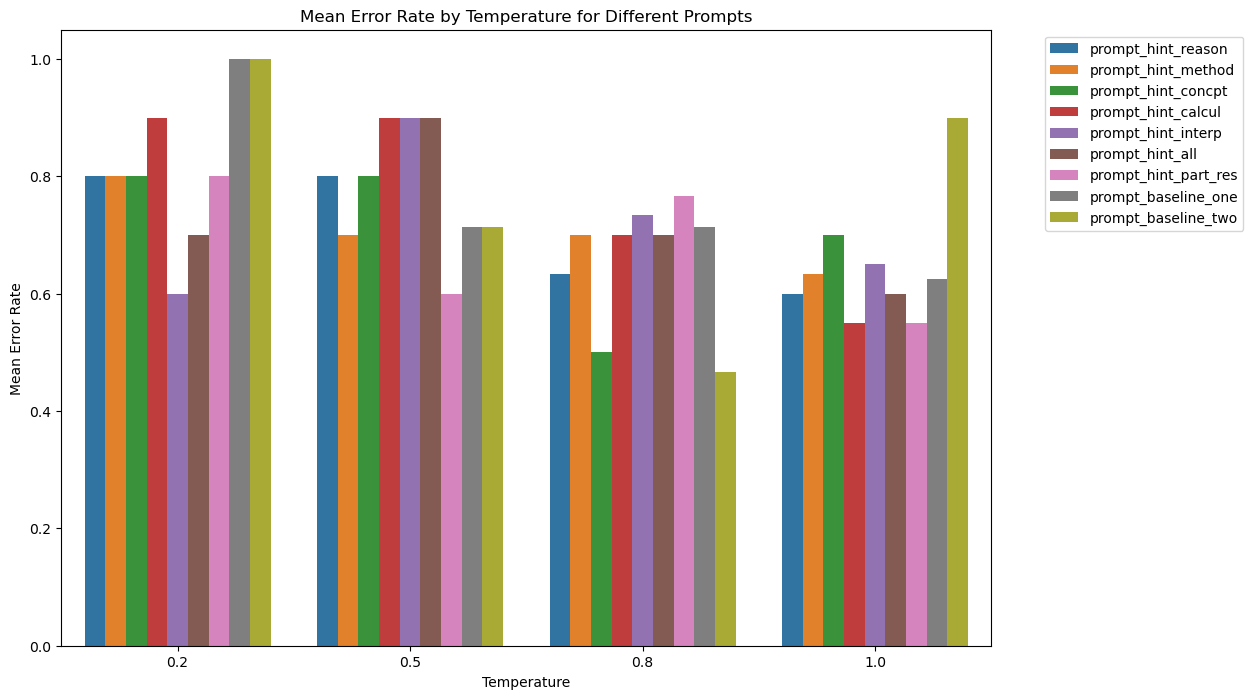}
        \label{fig:llama}
    }
    \hfill
    \subfigure[Mistral-7B-Instruct-v0.3]{
        \includegraphics[width=0.65\textwidth]{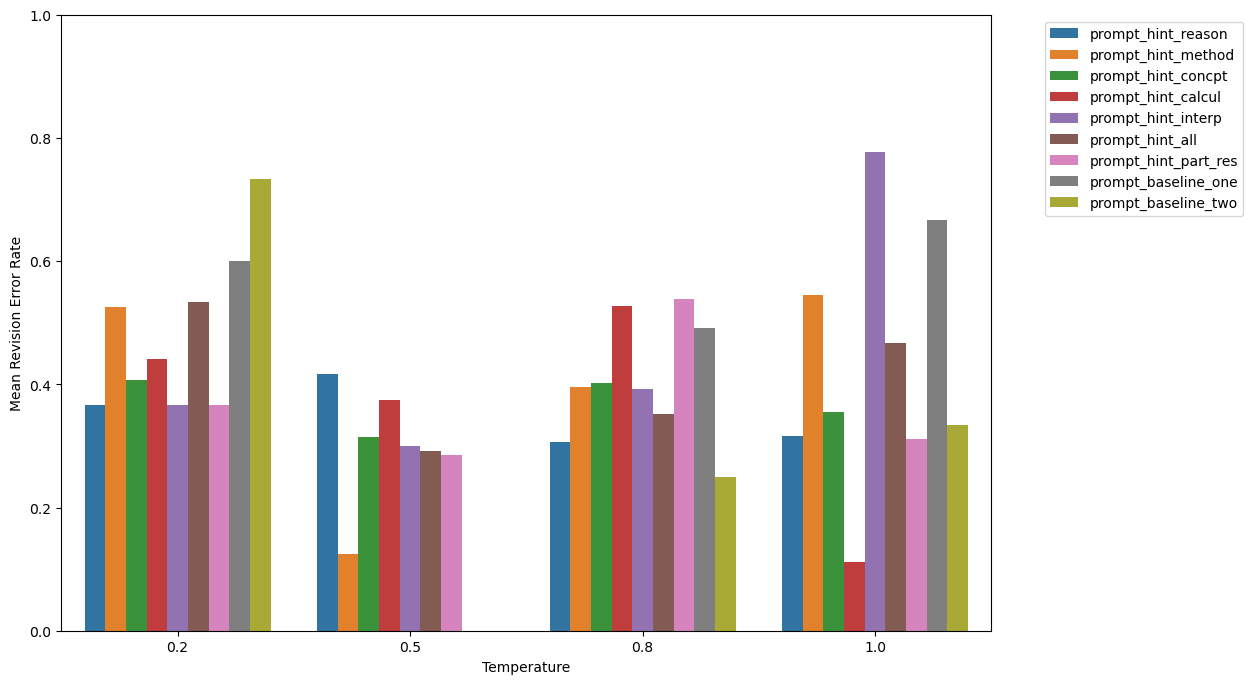}
        \label{fig:mistral}
    }
    \caption{Comparison of mean revision error rates across different temperatures and prompts for hint generation in three student models (GPT-3.5-turbo, Llama-3-8B-instruct, Mistral-7B-instruct-v3) \textbf{on exercise 1 - module 1}. The Figure illustrates how each model’s mean revision error rate evolves as the temperature increases and with varying prompts, highlighting how effectively the models can correct themselves using the provided hints. The accuracies before revision are as follows: For \textbf{GPT-3.5-turbo}, the accuracies are 0.00 at a temperature of 0.0, 0.05 at 0.2, 0.35 at 0.5, 0.282 at 0.8, and 0.4 at 1.0. For \textbf{Llama-3-8B-instruct}, the accuracies are 0.795 at 0.2, 0.636 at 0.5, 0.5 at 0.8, and 0.577 at 1.0. \textbf{Mistral-7B-instruct-v3} exhibits accuracies of 0.737 at 0.2, 0.795 at 0.5, 0.765 at 0.8, and 0.824 at 1.0.}
    \label{fig:mean_error_ex1mod1}
\end{figure}

The comparison is shown in Figure \ref{fig:mean_error_ex1mod1}, \ref{fig:mean_error_ex2mod2} and \ref{fig:mean_error_ex3mod7}.

\subsubsection{Additional results on the distribution of error revision rates by prompt and temperature for each student model across other exercises}
\label{sec:add_res_err}
\begin{figure}[htbp]
    \centering
    \subfigure[ GPT-3.5-turbo]{
        \includegraphics[width=0.47\textwidth]{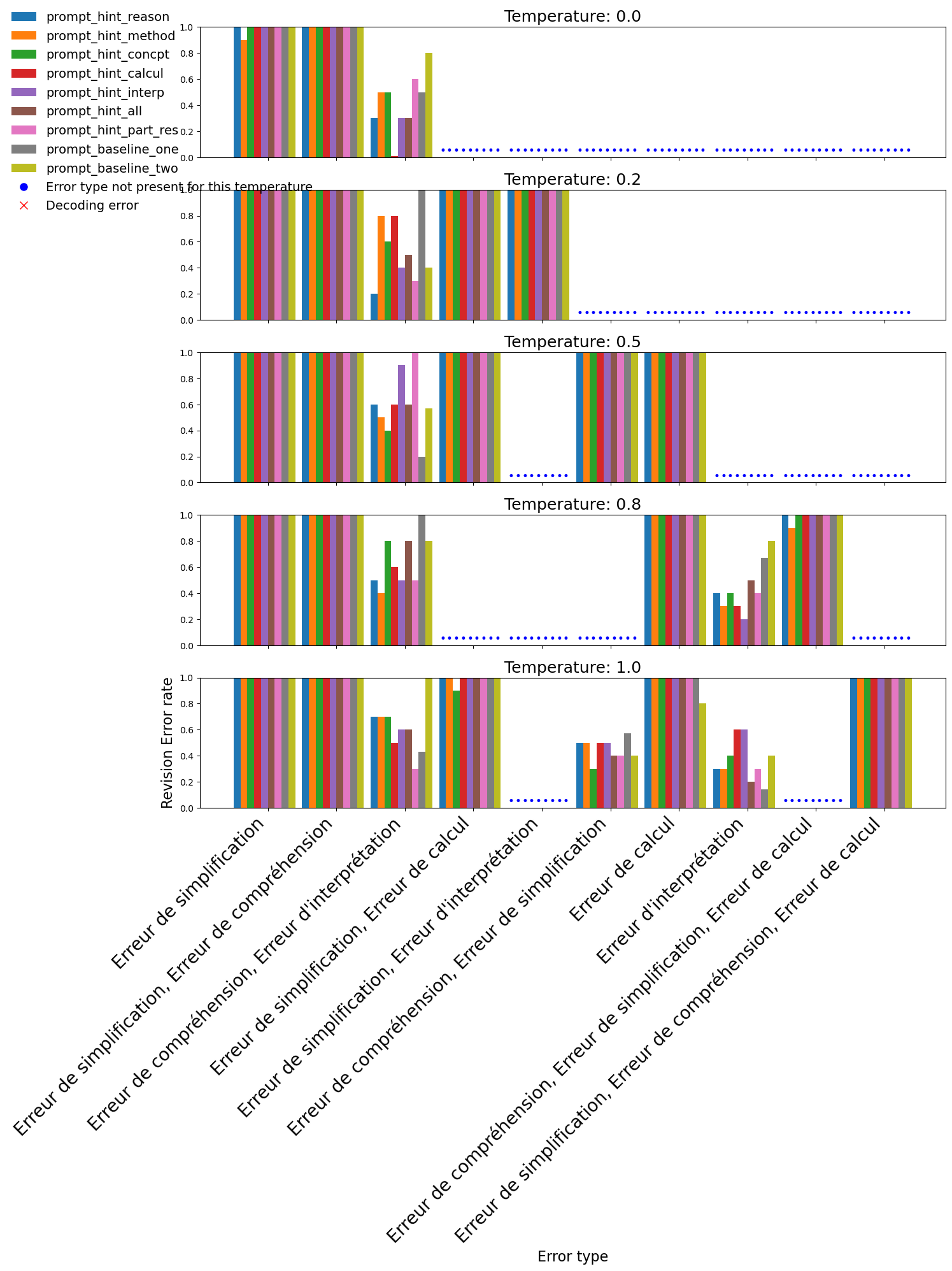}
    }
    \subfigure[Llama-3-8B-instruct]{
        \includegraphics[width=0.47\textwidth]{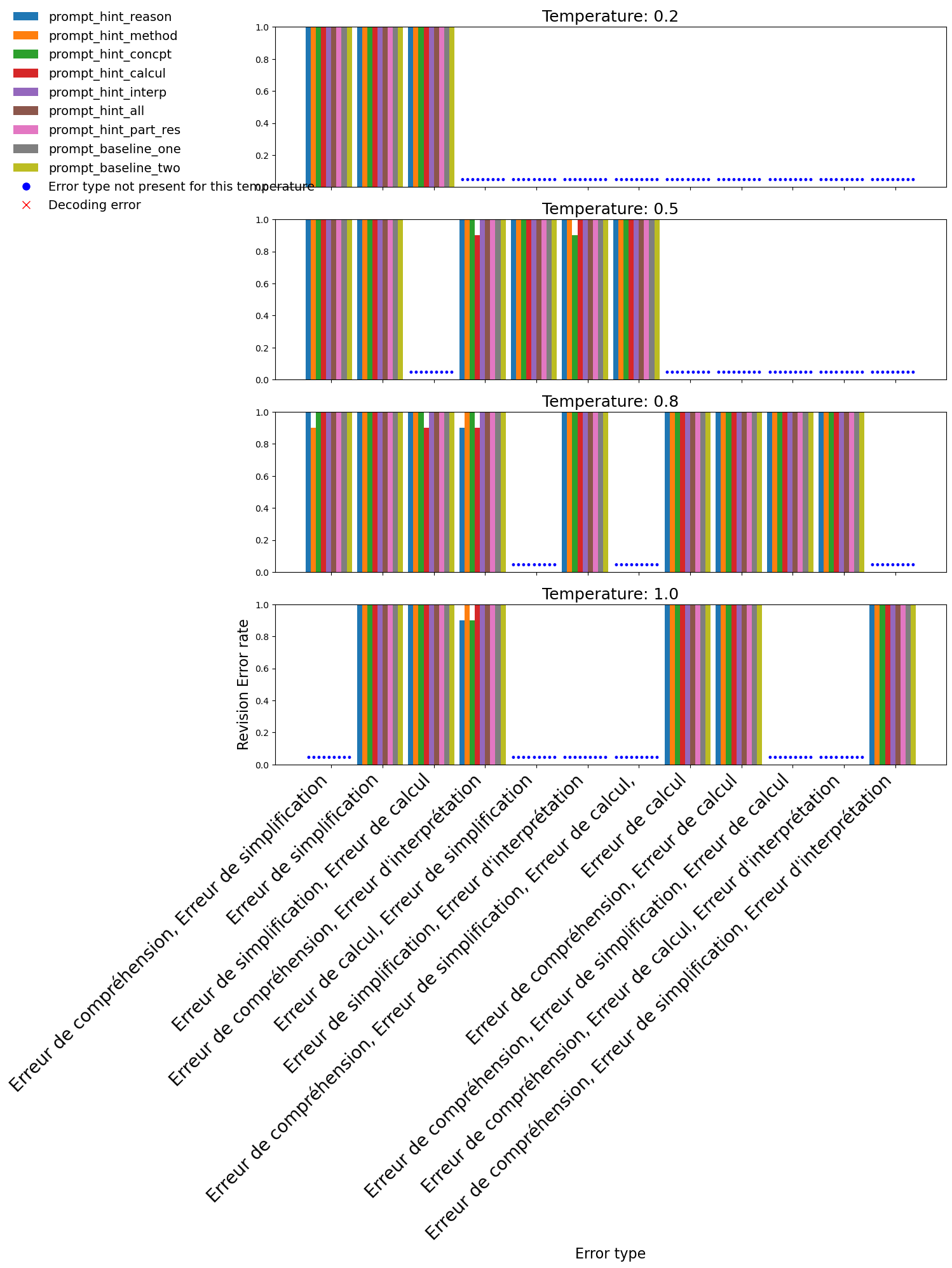}
    }
    \subfigure[Mistral-7B-Instruct-v0.3]{
        \includegraphics[width=0.47\textwidth]{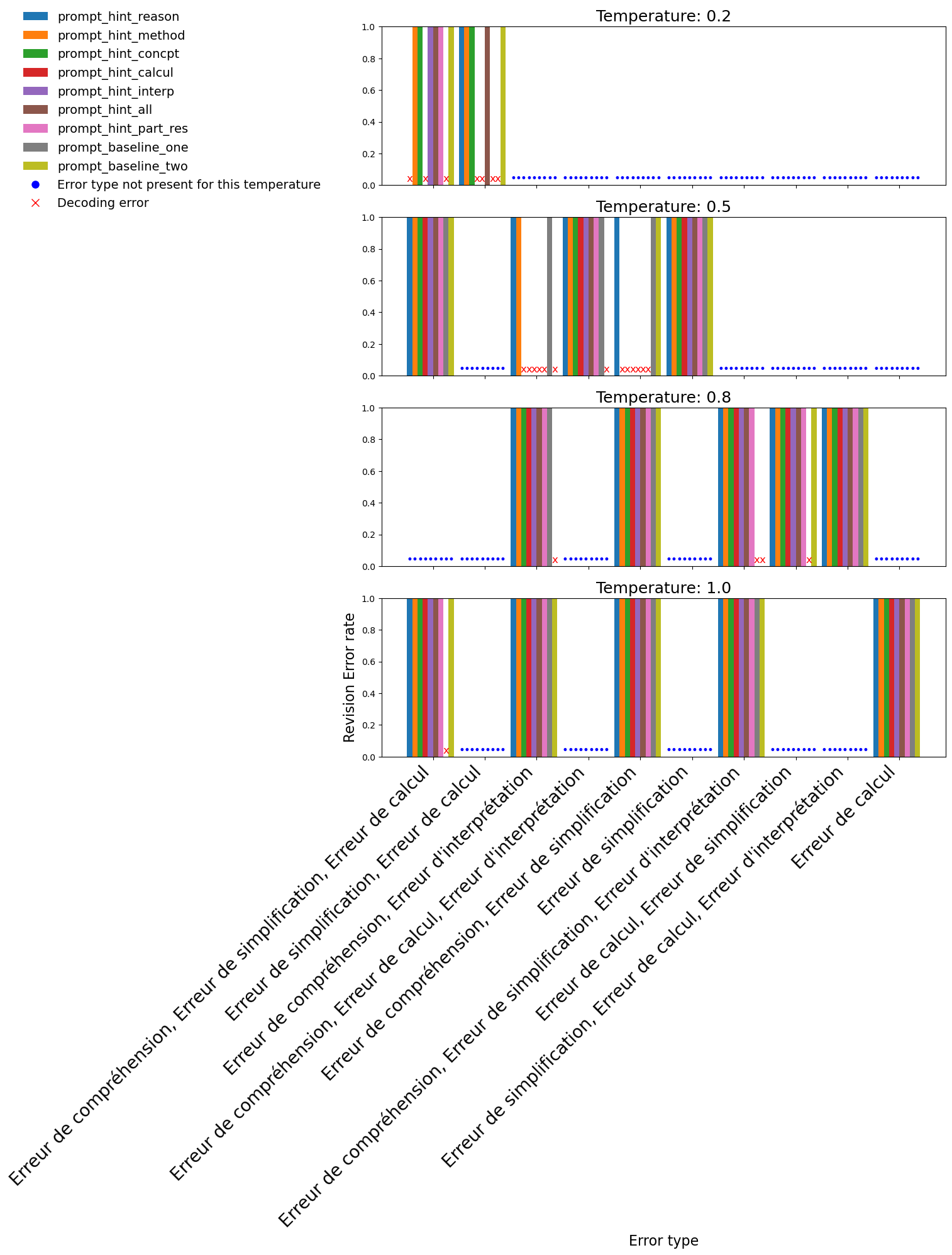}
    }
    \caption{Comparison of error revision rates across different temperatures and prompts for hint generation in three student models (GPT-3.5-turbo, Llama-3-8B-instruct, Mistral-7B-instruct-v3) on exercise 3 - module 7 with GPT-4o as teacher. The figure demonstrates that this exercise is challenging to solve, as indicated by the high error revision rate across all temperatures for all student models except GPT-3.5-Turbo, which attempts to self-correct using the hints.}
    \label{fig:error_revision_rates_ex3mod7}
\end{figure}

\begin{figure}[htbp]
    \centering
    \subfigure[GPT-3.5-turbo]{
        \includegraphics[width=0.47\textwidth]{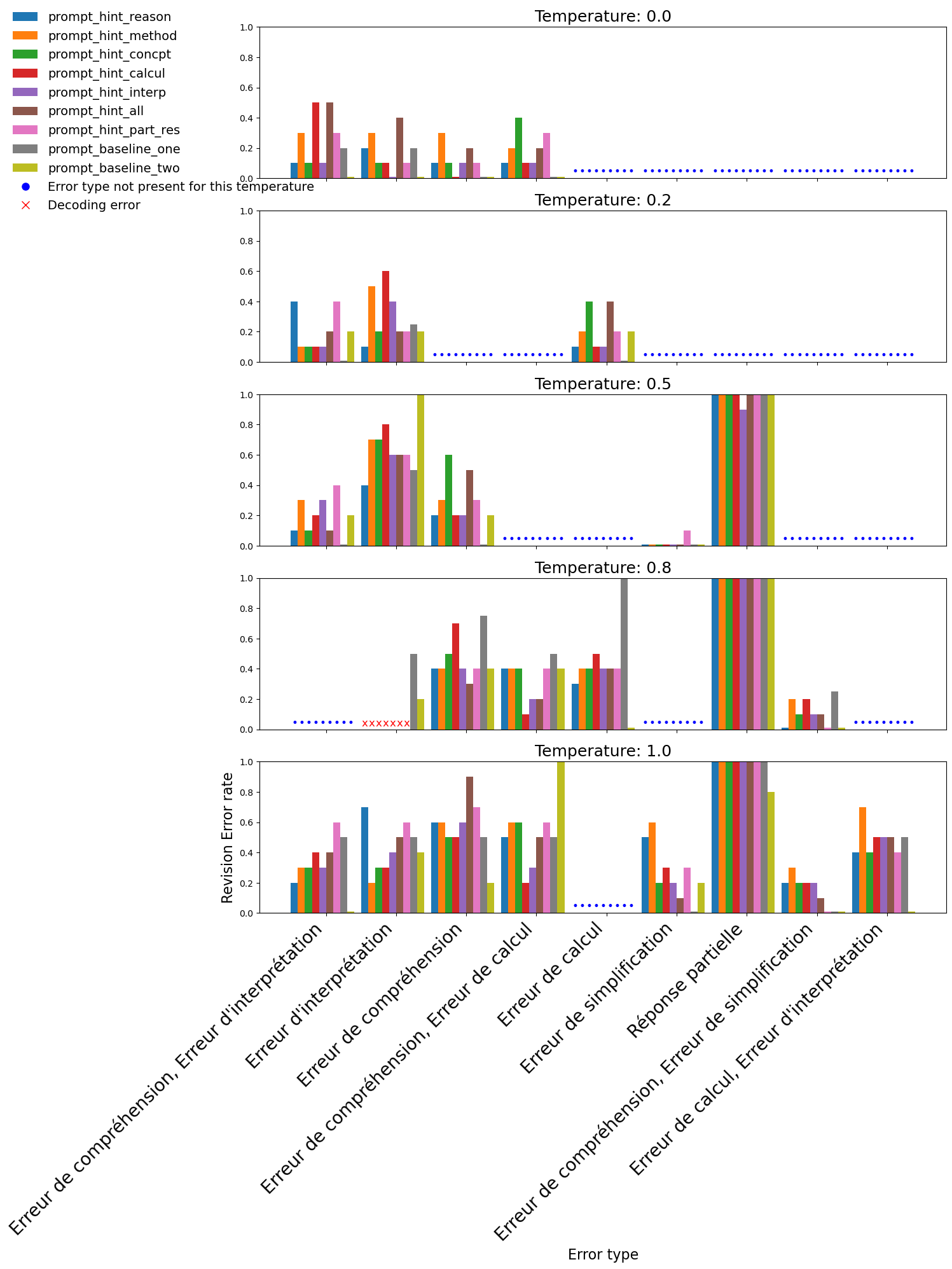}
    }
    \hfill
    \subfigure[Llama-3-8B-instruct]{
        \includegraphics[width=0.47\textwidth]{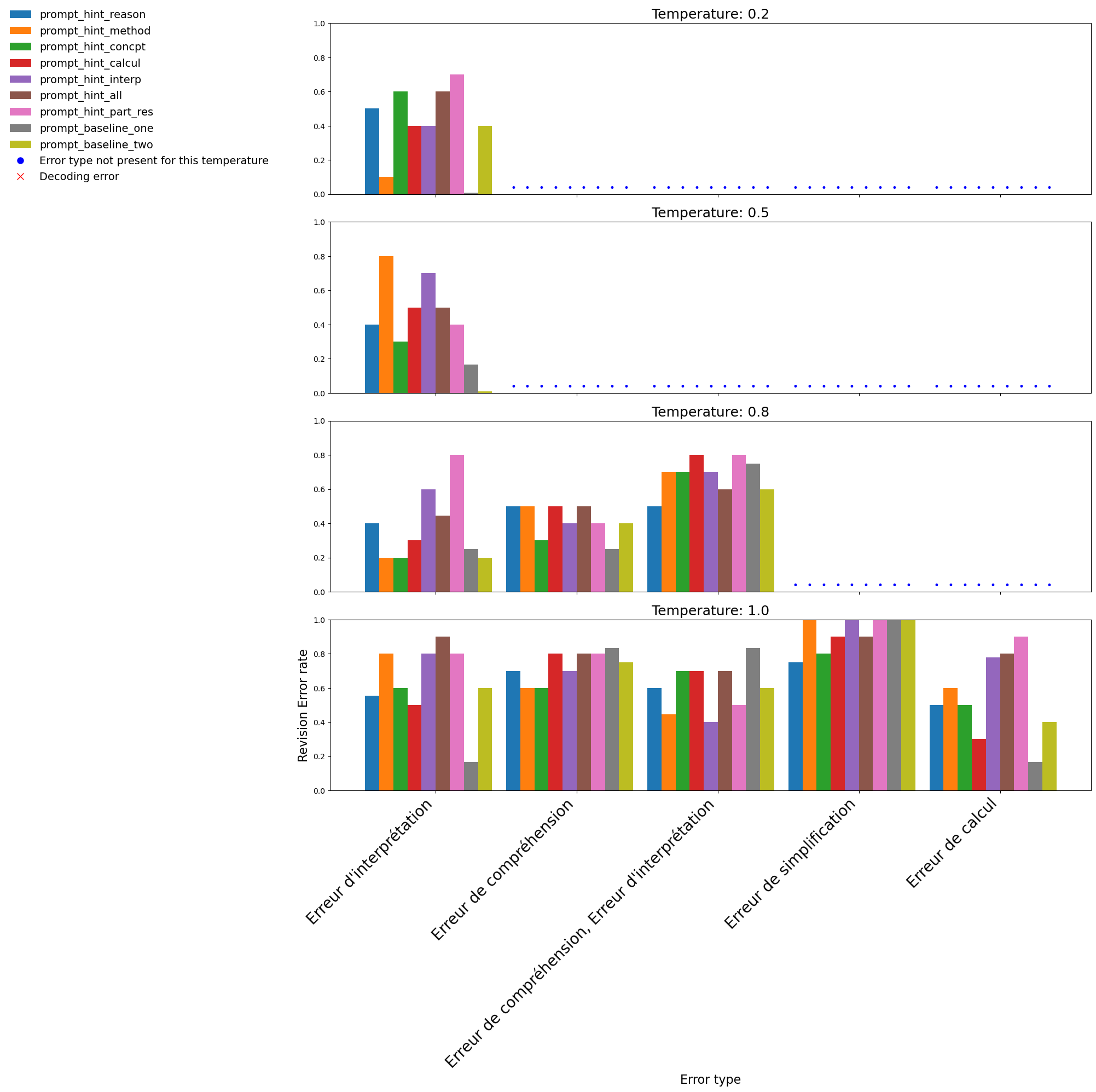}
    }
    \hfill
    \subfigure[Mistral-7B-Instruct-v0.3]{
        \includegraphics[width=0.47\textwidth]{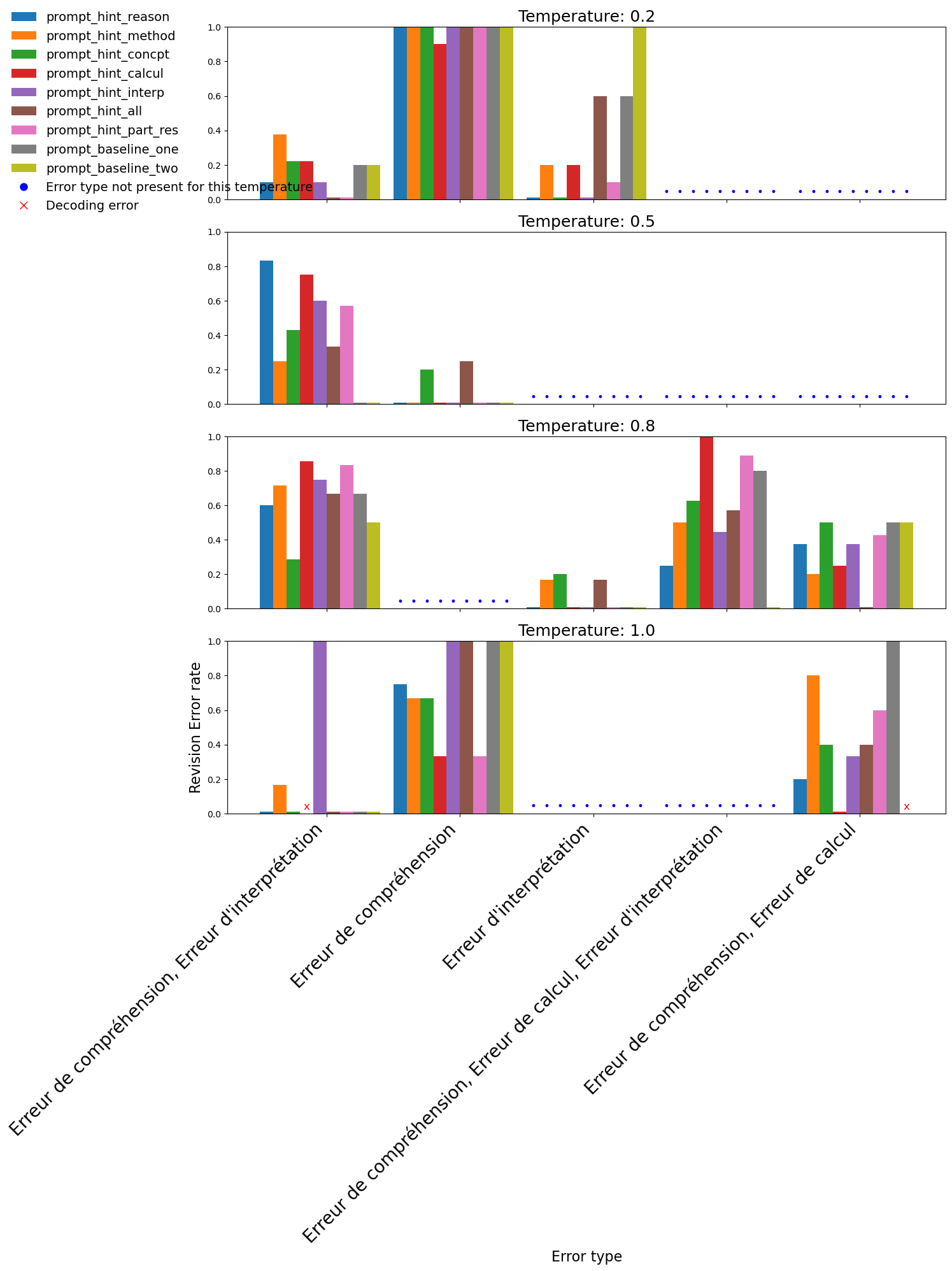}
    }
    \caption{Comparison of error revision rates across different temperatures and prompts for hint generation in three student models (GPT-3.5-turbo, Llama-3-8B-instruct, Mistral-7B-instruct-v3) on exercise 1-module 1. The Figure illustrates how each model’s error revision rate changes with increasing temperatures and with respect to types of errors, showing that the frequency of error types tends to increase at higher temperatures.}
    \label{fig:error_revision_rates_ex1mod1}
\end{figure}

\begin{figure}[htbp]
    \centering
    \subfigure[GPT-3.5-turbo]{
        \includegraphics[width=0.47\textwidth]{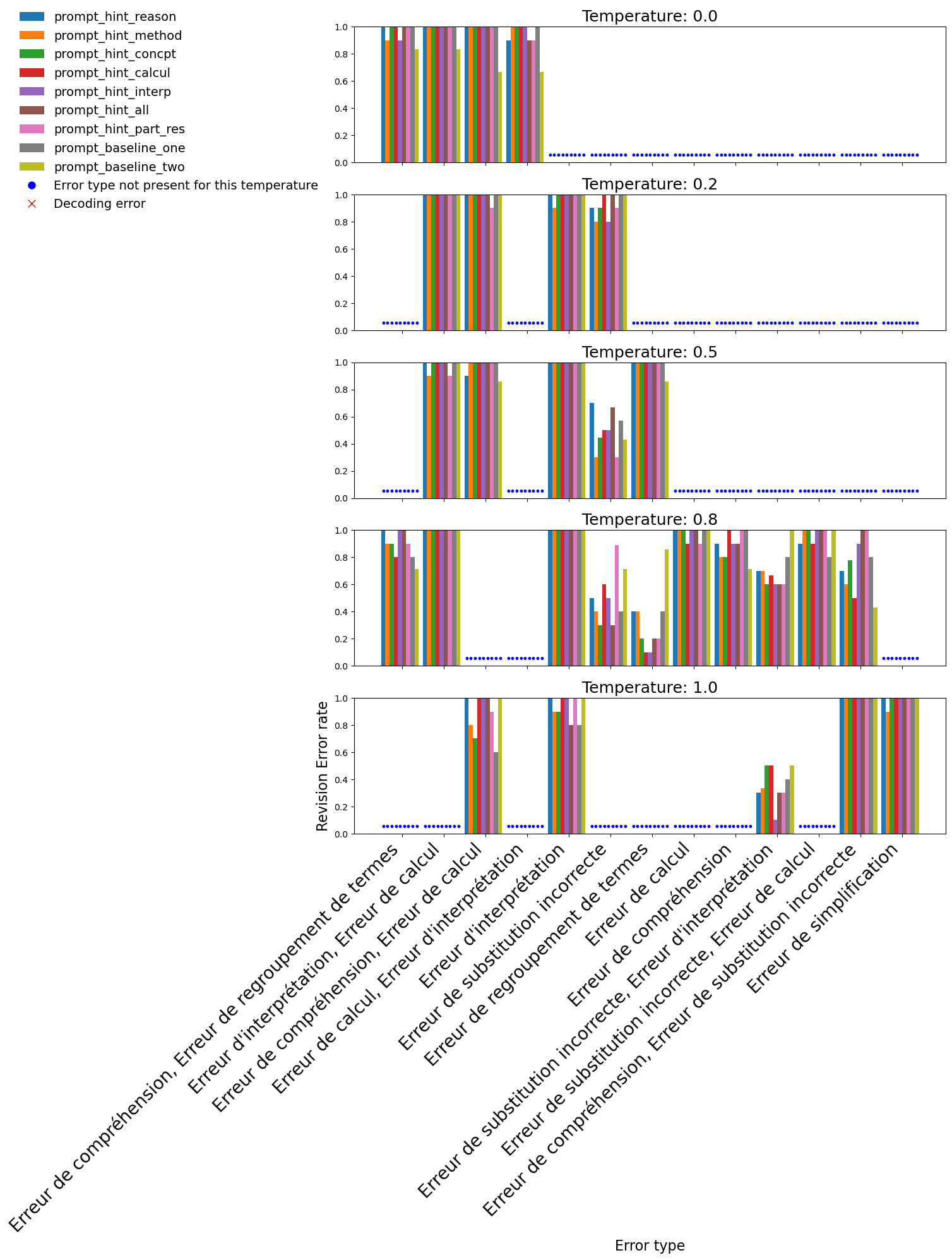}
        \label{fig:gpt_distrib}
    }
    \hfill
    \subfigure[Llama-3-8B-instruct]{
        \includegraphics[width=0.47\textwidth]{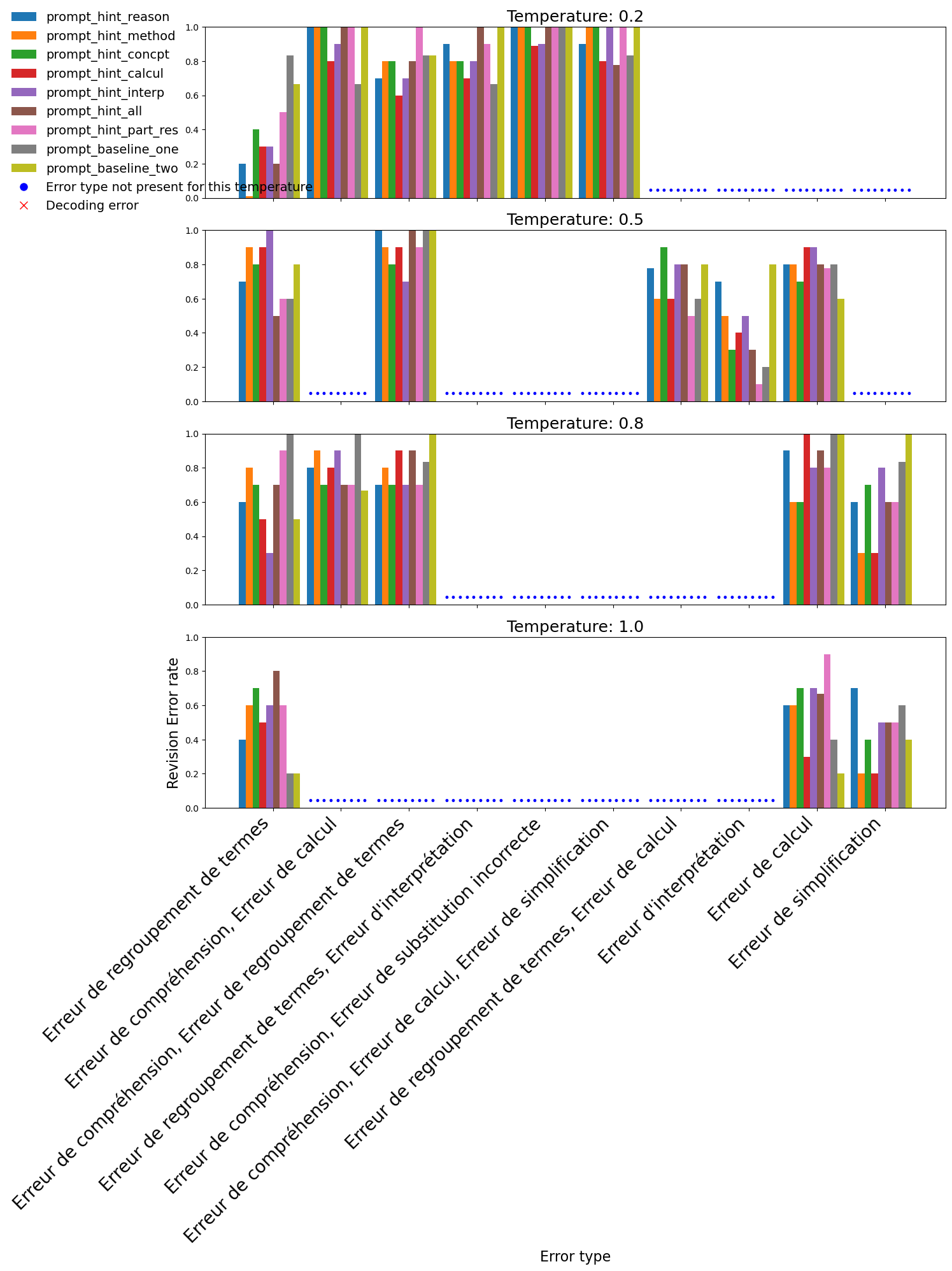}
        \label{fig:Llama_distrib}
    }
    \hfill
    \subfigure[Mistral-7B-Instruct-v0.3]{
        \includegraphics[width=0.47\textwidth]{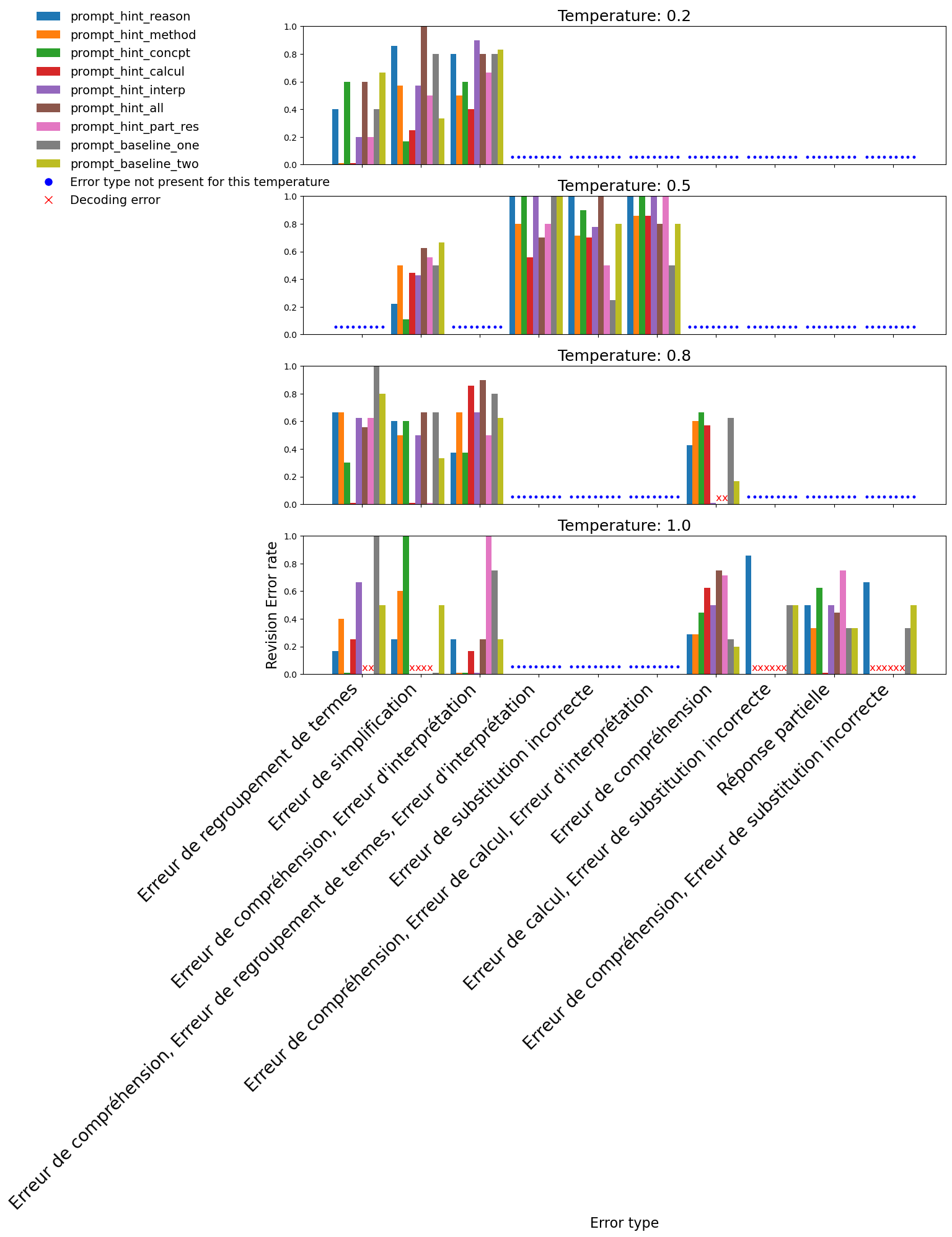}
        \label{fig:mistral_distrib}
    }
    \caption{Comparison of error revision rates across different temperatures and prompts for hint generation in three student models (GPT-3.5-turbo, Llama-3-8B-instruct, Mistral-7B-instruct-v3) on exercise 2 - module 2 with GPT-4o as teacher. The Figure illustrates how each model’s error revision rate changes with increasing temperatures and with respect to types of errors, showing that the frequency of error types tends to increase at higher temperatures.}
    \label{fig:error_revision_rates_ex2mod2}
\end{figure}

The comparison is shown in Figure \ref{fig:error_revision_rates_ex1mod1}, \ref{fig:error_revision_rates_ex2mod2} and \ref{fig:error_revision_rates_ex3mod7}.

\subsection{Description of exercises and pedagogical elements}
\label{sec:exercises}

This section presents the exercises used in our study, including key pedagogical elements such as the cognitive approach associated with each exercise, the type of exercise, the exercise statement, instructions, and the corresponding answer. The exercises were originally written in French, and the experiments were conducted in French as well. The prompts were also written in French. For the purpose of this paper, we have translated them into English.

\subsubsection{Exercise 1 - module 1}

\textbf{Cognitive Approach:} Transition from the concept of partitioning to the concept of fraction as a quotient, through the imposition of a constraint on the whole. \\
\textbf{Level 1} \\
This series of activities (A7, A8, A9) aims to gradually move beyond the intuitive notion that an equitable division of a whole composed of multiple units requires taking an equal part of each unit. Starting from Level 2, a condition imposed in the problem statement ``forces'' the student to counter this conception. The goal is to progressively reach the understanding of a fraction as a quotient. At Level 1, the statement aligns with the student’s intuitive conception, with no imposed conditions.

\textbf{Type of Exercise:} The student is presented with a problem involving the division of a whole, composed of $n$ units, into $m$ parts. \\
Example: The whole consists of 4 units, represented by 4 wooden planks. The whole is divided into 3 equal parts, and the student is asked to interpret the value of one part's size, given the condition of equal portions of each unit. The statement allows the student with a partition-based understanding of fractions to solve the problem by reasoning as follows: $\frac{4}{3}$ is like 4 times one-third of 1.

\textbf{Exercise Statement:} \\
``Elias bought two quiches of the same size. He decides to eat one-quarter of the quiches and save the rest for later. He wants to eat an equal part of each quiche. \textbf{What fraction of each quiche will he eat?}''

\textbf{Instruction:} \\
\textbf{Complete the following sentence with fractions:} Elias will eat \_\_\_\_\_ of the first quiche and \_\_\_\_\_ of the second quiche.

\textbf{Answer:} \\
Elias will eat $\frac{1}{4}$ of the first quiche and $\frac{1}{4}$ of the second quiche.

\subsubsection{Exercise 2 - module 2}

\textbf{Cognitive Approach:} Adopt a dual perspective to model a multi-step algebraic problem using a literal expression represented both as a sum and as a product. \\
\textbf{Level 4} \\
This series of activities (A1, A2, A3, A4) encourages flexibility in problem-solving strategies, moving beyond the strategy suggested by the problem's context and enabling the student to consider an alternative strategy based on the distributive property. This activity reinforces mastery of this property. Depending on the problem scenario, the student’s intuitive approach might involve modeling with a literal expression in either expanded form (sum of expressions) or factored form (product of expressions). \\
At Level 4, one step in the problem requires expressing one variable in terms of another, with an additional challenge introduced as the relationship between these two variables is expressed as a ratio. This ratio involves either multiplying by a fraction less than 1 or dividing by a whole number. For example, $\frac{1}{6}$ of a tulip corresponds to 1 rose.

\textbf{Type of Exercise:} The student is asked to model a two-step problem using a literal expression by selecting one or more correct answers from four given options.

\textbf{Exercise Statement:} \\
``To decorate her house, Julie enters a store and buys 5 of each of the following items: green plants and flower pots. The price of a green plant varies depending on the store's stock. A green plant costs 3 times as much as a matching flower pot. \\
Let $p$ be the price of a green plant. \textbf{How much did Julie pay in total?}''

\textbf{Instruction:} \\
\textbf{Identify the expressions that represent the total price Julie paid.} \\
Select the correct answer(s): $5p + \frac{5p}{3}$? $5(p + \frac{p}{3})$? $5p + \frac{p}{3}$?

\textbf{Answer:} \\
$5p + \frac{5p}{3}$ and $5(p + \frac{p}{3})$.

\subsubsection{Exercise 1 - module 2 (Similar to the previous module, only the exercise is presented here)}

\textbf{Exercise Statement:} \\
``In a restaurant, there are 30 tables. Each table has a bouquet. The number of roses in each bouquet varies depending on the arrivals. Each bouquet contains three times fewer roses than tulips. \\
Let $r$ be the number of roses. \textbf{How many flowers are there in total in this restaurant?}''

\textbf{Instruction:} \\
\textbf{Identify the expressions that represent the total number of flowers in the restaurant.} \\
Select the correct answer(s): $30(3r + r)$? $120r$? $30(\frac{r}{3} + r)$?

\textbf{Answer:} \\
$30(3r + r)$ and $120r$.

\subsubsection{Exercise 3 - module 7}

\textbf{Cognitive Approach:} Understand how to simplify a fraction to its irreducible form. \\
\textbf{Level 1} \\
This series of activities (A3, A4) focuses on using prime factorization to determine whether a fraction is in its simplest form. It aims to develop conceptual expertise in fractional operations and explore the different meanings of fractions in problem-solving contexts. At Level 1, the scenarios align with the intuitive understanding of a fraction as a ratio between a part and a whole.

\textbf{Type of Exercise:} The student answers (yes or no) a question regarding the simplifiability of the fraction presented in the problem. If the fraction is reducible, the student writes its simplified form and completes a response sentence.

\textbf{Exercise Statement:} \\
``A truck driver covered five thousand five hundred thirty-thirds of a kilometer in two hours.''

\textbf{Instruction:} \\
\textbf{Can the fraction that describes the number of kilometers the truck driver covered be simplified?} Yes? No? If yes: If the truck driver covered five thousand five hundred thirty-thirds of a kilometer in two hours, that means he covered \_\_\_\_\_ kilometers in two hours.

\textbf{Answer:} \\
The correct answer is yes. If the truck driver covered five thousand five hundred thirty-thirds of a kilometer in two hours, that means he covered $\frac{500}{3}$ kilometers in two hours.

\subsection{Example of good and poor revision of student models on exercise 1 - module 1 after receiving a hint from the teacher model using one of the best prompts}
The conversation between teacher and student is in French in our experiments, but we have translated it into English for the purposes of this paper.
\label{sec:example_conversation}
\begin{itemize}
    \item \textbf{Teacher Model: Llama-3-8B-instruct}
    \begin{itemize}
        \item \textbf{Student Model: GPT-3.5-turbo}
        \begin{description}
            \item[Initial Reasoning:] Elias decided to eat a quarter of both quiches, so he will eat 1/4 of each quiche. Since he wants to eat an equal part of each quiche, he will eat half of the portion he took from each quiche, which is 1/2 * 1/4 = 1/8 of each quiche.
            \item[Initial Answer:] Elias will eat 1/8 of the first quiche and 1/8 of the second quiche.
            \item[Teacher Model Hint:] How can you use the notion of a quarter to break down the portion Elias eats from each quiche?
            \item[Revised Answer (True):] Elias will eat 1/4 of the first quiche and 1/4 of the second quiche.
            \item[Teacher Model Hint:] Can you explain why it is reasonable to think that Elias will eat the same portion of each quiche, even if it means he doesn’t eat the entire quarter of the quiches?
            \item[Revised Answer (False):] Elias will eat 1/8 of the first quiche and 1/8 of the second quiche.
        \end{description}
        
        \item \textbf{Student Model: Mistral-7B-v0.3}
        \begin{description}
            \item[Initial Reasoning:] Elias decided to eat a quarter of each quiche. So he ate half of the first quiche and half of the second quiche.
            \item[Initial Answer:] Elias will eat 1/2 of the first quiche and 1/2 of the second quiche.
            \item[Teacher Model Hint:] What does it mean if Elias decided to eat a quarter of the quiches, not an equal part of each quiche?
            \item[Revised Answer (True):] Elias will eat 1/4 of the first quiche and 1/4 of the second quiche.
            \item[Teacher Model Hint:] Can you explain why the 1/4 answer is justified by the phrase "He decides to eat a quarter of the quiches"?
            \item[Revised Answer (False):] Elias will eat 1/2 of the first quiche and 1/2 of the second quiche.
        \end{description}
        
        \item \textbf{Student Model: Llama-3-8B-instruct}
        \begin{description}
            \item[Initial Reasoning:] Elias bought two quiches of the same size and decided to eat a quarter of the quiches. This means he will eat 1/4 of the total quiches. To eat an equal part of each quiche, he will eat 1/4 / 2 = 1/8 of each quiche.
            \item[Initial Answer:] "First quiche": "1/8" \newline "Second quiche": "1/8"
            \item[Teacher Model Hint:] Can you explain why you divided a quarter by two to find the portion of each quiche that you will eat?
            \item[Revised Answer (True):] Elias will eat 1/4 of the first quiche and 1/4 of the second quiche.
            \item[Teacher Model Hint:] How can we think of sharing the quiches as an equitable distribution of the entire quiches, even if the portions are not equal?
            \item[Revised Answer (False):] Elias will eat 1/2 of the first quiche and 1/2 of the second quiche.
        \end{description}
    \end{itemize}

    \item \textbf{Teacher Model: GPT-4o}
    \begin{itemize}
        \item \textbf{Student Model: GPT-3.5-turbo}
        \begin{description}
            \item[Initial Reasoning:] Elias decided to eat a quarter of both quiches, so he will eat 1/4 of each quiche. Since he wants to eat an equal part of each quiche, he will eat half of the portion he took from each quiche, which is 1/2 * 1/4 = 1/8 of each quiche.
            \item[Initial Answer:] Elias will eat 1/8 of the first quiche and 1/8 of the second quiche.
            \item[Teacher Model Hint:] How can you divide what he will eat into equal parts coming from two identical quiches?
            \item[Revised Answer (True):] Elias will eat 1/4 of the first quiche and 1/4 of the second quiche.
            \item[Teacher Model Hint:] If Elias eats a quarter of the quiches, what fraction of each quiche does that represent?
            \item[Revised Answer (False):] Elias will eat 1/8 of the first quiche and 1/8 of the second quiche.
        \end{description}

        \item \textbf{Student Model: Llama-3-8B-instruct}
        \begin{description}
            \item[Initial Reasoning:] Elias bought two quiches of the same size and decided to eat a quarter of the quiches. This means he will eat 1/4 of the total quiches. To eat an equal part of each quiche, he will eat 1/4 / 2 = 1/8 of each quiche.
            \item[Initial Answer:] "First quiche": "1/8" \newline "Second quiche": "1/8"
            \item[Teacher Model Hint:] What fraction of each quiche should he eat if he wants the sum of the fractions eaten from both quiches to equal a quarter of the total quiches?
            \item[Revised Answer (True):] Elias will eat 1/4 of the first quiche and 1/4 of the second quiche.
            \item[Teacher Model Hint:] Elias eats an equal part of each quiche and consumes a quarter in total. How can you divide the total consumption for each quiche?
            \item[Revised Answer (False):] Elias will eat 1/8 of the first quiche and 1/8 of the second quiche.
        \end{description}

        \item \textbf{Student Model: Mistral-7B-v0.3}
        \begin{description}
            \item[Initial Reasoning:] Elias decided to eat a quarter of each quiche. So he ate half of the first quiche and half of the second quiche.
            \item[Initial Answer:] Elias will eat 1/2 of the first quiche and 1/2 of the second quiche.
            \item[Teacher Model Hint:] How can Elias divide the total amount he wants to eat, which is a quarter of all the quiches, equally between the two quiches?
            \item[Revised Answer (True):] Elias will eat 1/4 of the first quiche and 1/4 of the second quiche.
            \item[Teacher Model Hint:] What would the calculation be if Elias decided to eat a quarter of the total amount of the two quiches together rather than eating a quarter of each quiche individually?
            \item[Revised Answer (False):] Elias will eat 1/2 of the total amount of the two quiches.
        \end{description}
    \end{itemize}
\end{itemize}

\end{document}